\documentclass[conference]{IEEEtran}
\IEEEoverridecommandlockouts
\usepackage{cite}
\usepackage{amsmath,amssymb,amsfonts}
\usepackage{algorithmic}
\usepackage{algorithm}
\usepackage{graphicx}
\usepackage{textcomp}
\usepackage{xcolor}

\definecolor{lightgray}{RGB}{245,245,245}
\definecolor{blockgray}{RGB}{235,240,248}
\definecolor{oursblue}{RGB}{215,225,245}
\usepackage{colortbl}
\usepackage{booktabs}
\usepackage{multirow}

\def\BibTeX{{\rm B\kern-.05em{\sc i\kern-.025em b}\kern-.08em
    T\kern-.1667em\lower.7ex\hbox{E}\kern-.125emX}}
\begin{document}

\title{Operator Learning with Spatial-Temporal Disentanglement for Multi-Scale Long-Term Dynamics Prediction
\thanks{*These authors contributed equally to this work.}
\thanks{†Zhenzhong Wang is the corresponding authors.}
}

\author{\IEEEauthorblockN{\qquad \qquad \ \ \ Chunhe Liu*\ \ \qquad \qquad \ }
\IEEEauthorblockA{\textit{The University of Hong Kong \&} \\
\textit{Xiamen University}\\
Hong Kong, China \\
u3637314@connect.hku.hk}
\and
\IEEEauthorblockN{\qquad \qquad \ \ \ Yining Qin*\ \ \qquad \qquad \ }
\IEEEauthorblockA{\textit{Xiamen University Malaysia} \\
Sepang, Malaysia \\
CST2409043@xmu.edu.my}
\and
\IEEEauthorblockN{\qquad \qquad \ \ \ Yipeng Huang\ \qquad \qquad \ }
\IEEEauthorblockA{\textit{Xiamen University} \\
Xiamen, China \\
yipenghuang@stu.xmu.edu.cn }
\and
\IEEEauthorblockN{\qquad \qquad \ \ \ \ Xin Zhang\ \ \ \qquad \qquad \ }
\IEEEauthorblockA{\textit{Xiamen University} \\
Xiamen, China \\
sidxinz@stu.xmu.edu.cn}
\and
\IEEEauthorblockN{\qquad \qquad \ Zhenzhong Wang†\qquad \qquad \ }
\IEEEauthorblockA{\textit{Xiamen University} \\
Xiamen, China \\
zhenzhongwang@xmu.edu.cn}
\and
\IEEEauthorblockN{\qquad \qquad \ \ \ \ Min Jiang\ \ \qquad \qquad \ }
\IEEEauthorblockA{\textit{Xiamen University} \\
Xiamen, China \\
minjiang@xmu.edu.cn}
}

\maketitle

\begin{abstract}
Learning operators for time-evolving physical fields is central to modeling complex dynamical systems. Existing neural operators typically encode spatial and temporal information into a shared latent space, where the two components are entangled, hindering the explicit modeling of multiscale physical details and temporal dynamics. To address this, we explicitly disentangle the time-evolving fields into separate temporal and spatial components at multiple scales, rather than compressing them into a unified latent space. Specifically, we apply a 2-D wavelet decomposition along the spatial dimensions to capture global physical structures as well as fine-grained spatial details, and a 1-D Haar wavelet decomposition along the temporal dimension to model temporal evolution trends. Building upon this decomposition, we further devise a Fourier branch block to capture long-range dependencies across both space and time.
On representative multi-physics benchmarks, our approach not only outperforms state-of-the-art neural operators in accuracy and robustness, but also maintains reliable performance under limited data, noisy inputs, and super-resolution evaluation.
\end{abstract}


\section{Introduction}
Learning the evolution dynamics of multiple coupled physical fields is a core problem across scientific computing, such as climate modeling, fluid–structure interaction, and materials science~\cite{Wong2025EvolutionaryOO,kd1,kd2,kd3,11351701}. Traditionally, these dynamics are governed by coupled partial differential equations (PDEs) and solved via numerical discretization \cite{szj2,szj1}. While these computational solvers provide high-fidelity approximations, they are computationally expensive \cite{tc,tc1}.

\begin{figure}[t]
    \centering
    \includegraphics[width=1\linewidth]{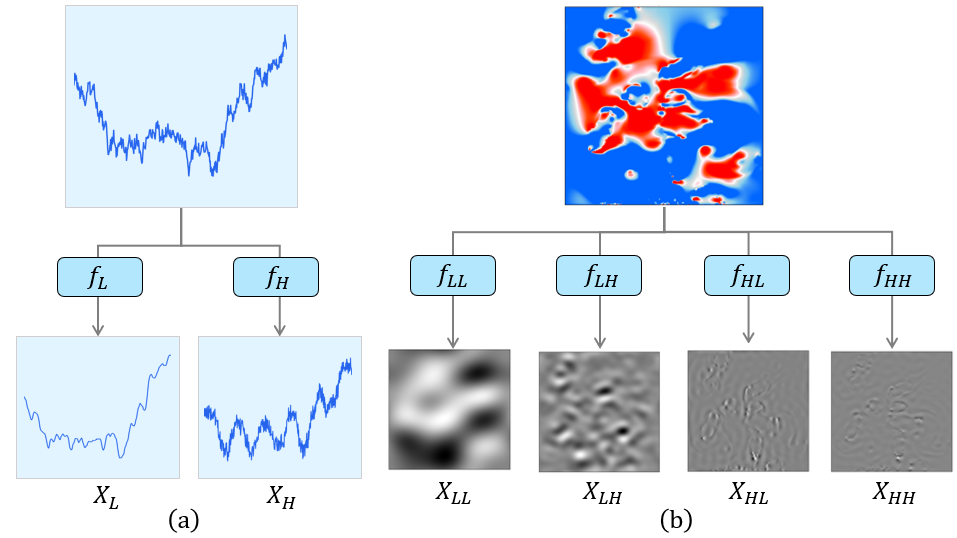}
        \caption{(a) 1D Haar decomposition along the time dimension captures long-term evolution trends $X_L$ and local rapid fluctuations $X_H$. (b) 2D Haar transform decomposes the spatial physical field into a low-frequency component $X_{LL}$ and three high-frequency components $X_{LH}$, $X_{HL}$, and $X_{HH}$, corresponding to approximation, horizontal, vertical, and diagonal details, respectively, thereby capturing spatial features at different frequencies.}
        \vspace{-0.3cm}
    \label{fig:intro}
\end{figure}

Recently, data-driven surrogate models have emerged as efficient alternatives \cite{datad, piml,datad2,cvpr1,cvpr2}. Among these approaches, neural operators have shown the potential for learning mappings between parametric inputs and solution fields, enabling efficient prediction of evolution dynamics of physical fields~\cite{kovachki2023neural,cvpr3,cvpr4,bo,cross}. By directly approximating solution operators rather than discretized computational solvers, neural operator architectures such as the Fourier Neural Operator (FNO)~\cite{fno}, DeepONet ~\cite{deeponet,fvd}, and their extensions ~\cite{li2020gno,li2020gno2,qiu2024derivative, difffno} have demonstrated improved generalization across parametric families of PDEs.

However, most existing neural operators encode spatio-temporal dynamics within a unified latent space, applying identical spectral, convolutional, or attention mechanisms across spatio-temporal domains \cite{piml,datad2,sury}. This design implicitly assumes that the spatial and temporal components of the dynamics are isotropic \cite{st,st1}. For instance, along the spatial dimensions, the FNO performs a global spectral convolution on the whole fields~\cite{fno}. While effective for capturing smooth, large-scale structures, this operation inherently suppresses high-frequency components that are crucial for representing sharp spatial variations such as boundaries and localized patterns, leading to a well-known spectral bias toward low-frequency features~\cite{wangg2026work, ppc}. On the other hand, when temporal components are entangled with spatial representations in the same latent space, the model lacks an explicit mechanism to capture sequential temporal trends, which often results in temporal instability and degraded long-term prediction performance~\cite{peg, terr}.

To explicitly model multiscale physical details and temporal dynamics, we propose a neural operator that disentangles spatio-temporal coupling via Haar wavelet decomposition, which provides a principled mechanism for separating signals across frequency bands and yielding representations that respect the intrinsic structures of both domains.
As illustrated in Fig.~\ref{fig:intro}, along the temporal dimension, a 1-D Haar transform decomposes time-series signals into slowly varying evolution trends and rapidly changing temporal fluctuations. High-frequency temporal components, which are most susceptible to error accumulation during autoregressive rollout, are explicitly separated from low-frequency components for modeling \cite{wdno}. Along the spatial dimensions, a 2-D Haar-based transform decomposes each field into multiple frequency subbands (low-frequency component $X_{LL}$ and three high-frequency components $X_{LH}$, $X_{HL}$, and $X_{HH}$, which correspond to the approximation information, horizontal details, vertical details, and diagonal details) 
separating global low-frequency structures from localized high-frequency patterns \cite{wt}. This prevents dominant low-frequency modes from overwhelming spatially localized dynamics and alleviates the spectral bias.
Built upon this decomposition, the proposed Haar wavelet-based neural operator, termed Wavelet Decomposition Neural Operator (WDNO), processes spatial and temporal components separately, enabling dedicated modeling of multiscale physical structures and temporal evolution patterns. The key contributions are summarized as follows:

\begin{enumerate}
    \item We design a Temporal Wavelet Convolution (TWC) module to extract temporal features. TWC module integrates 1-D wavelet transform with convolution operations, decomposing the time series into slow-varying evolution trends and fast fluctuations, thereby mitigating temporal error accumulation and improving long-horizon prediction stability.

     \item We design a Spatial Wavelet Convolution (SWC) module for spatial feature extraction. The SWC module combines a 2-D wavelet transform with pointwise convolution, decomposing spatial representations into distinct frequency subbands. This design prevents global low-frequency modes from dominating the learning process and enables the operator to capture localized, non-stationary spatial patterns that are often underrepresented.

\item  We propose a Fourier branch block that captures long-range spatio-temporal dependencies in disentangled feature representations. By applying one-dimensional Fourier convolution along the temporal axis and two-dimensional Fourier convolution along spatial axes in parallel branches, the block efficiently models global correlations while preserving high-frequency local details from the wavelet-based decomposition.

 \item Extensive experiments on representative multi-physics benchmarks demonstrate that our method consistently outperforms state-of-the-art neural operators in terms of accuracy and robustness, while exhibiting strong generalization under low-data, noisy, and super-resolution settings.

\end{enumerate}
\section{Preliminaries}
Wavelet decomposition offers a principled multi-resolution analysis by separating a signal into low-frequency trends and high-frequency variations. 
In this work, we employ Haar-based 1-D and 2-D wavelet decompositions. While both share the same basis function, they correspond to fundamentally different assumptions about the underlying data topology: 1-D decomposition is appropriate for temporal sequences, whereas 2-D decomposition is tailored for spatial fields.

Consider a discrete signal $\mathbf{x} = [x_1, x_2, \dots, x_N]^\top \in \mathbb{R}^N$, where $N$ is even. The 1-D Haar transform operates on adjacent pairs of elements and produces approximation (low-frequency) and detail (high-frequency) coefficients as $a_k = \frac{x_{2k-1} + x_{2k}}{2}$, $d_k = \frac{x_{2k-1} - x_{2k}}{2}$, $k = 1, \dots, \frac{N}{2}$.
The same procedure can be recursively applied to the sequence of approximation coefficients $\{a_k\}$ to obtain a multi-level decomposition. This construction implicitly assumes that the signal possesses a temporal sequence, where adjacency is defined purely by ordering. Consequently, the detail coefficients measure local variations along a single direction, making the 1-D Haar decomposition particularly suitable for temporal signals, sequential measurements, and any data whose neighborhood relationships arise from sequence order rather than spatial data.

For a two-dimensional signal $\mathbf{X} \in \mathbb{R}^{H \times W}$ defined on a grid, where both $H$ and $W$ are even. The 2-D wavelet transform operates on each non-overlapping $3 \times 3$ block
\[
\begin{bmatrix}
x_{11} & x_{12} & x_{13}\\
x_{21} & x_{22} & x_{23}\\
x_{31} & x_{32} & x_{33}\\
\end{bmatrix},
\]
from which four coefficients are computed. The approximation coefficient is given by $a = \frac{x_{11} + x_{12} + x_{13} + x_{21} + x_{22} + x_{23} + x_{31} + x_{32} + x_{33}}{9}$,
while the horizontal, vertical, and diagonal detail coefficients are defined as $d^{(h)} = \frac{x_{11} - x_{13} + x_{21} - x_{23} + x_{31} - x_{33}}{3}$, $d^{(v)} = \frac{x_{31} + x_{32} + x_{33} - x_{11} - x_{12} - x_{13}}{3}$, and $d^{(d)} = \frac{x_{13} - x_{11} - x_{33} + x_{31}}{3}$.
The approximation component can again be recursively decomposed to obtain multi-resolution representations. Unlike the 1-D case, this construction assumes a grid topology, where each element has meaningful spatial neighbors in horizontal and vertical directions. The detail coefficients, therefore, encode local spatial structures such as edges, textures, and boundaries.
Therefore, the distinction between 1-D and 2-D wavelet decompositions is topological rather than merely dimensional. Despite these differences, both methods share a fundamental multi-resolution property: the ability to separate coarse trends from fine-scale variations. In this framework, approximation coefficients capture global structures, while detail coefficients isolate localized variations. We leverage this multi-scale property by applying 1-D Haar decomposition to sequence signals and 2-D decomposition to grid-structured field data.
\section{Proposed Method}
\begin{figure*}[t]
  \centering
\includegraphics[width=0.941\linewidth]{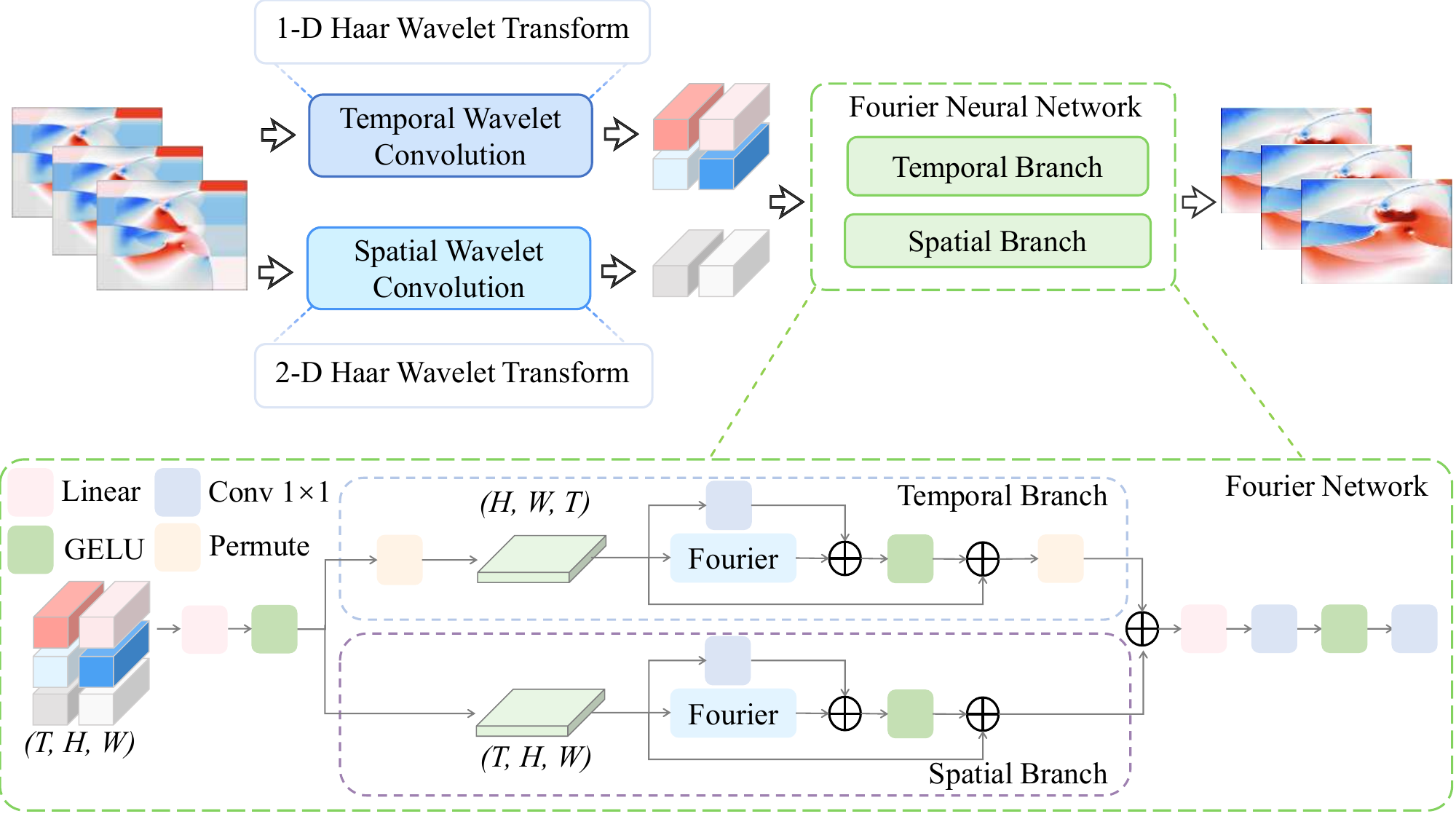}
  \caption{Schematic of WDNO architecture shows that Temporal Wavelet Convolution and Spatial Wavelet Convolution capture temporal and spatial features through 1D and 2D wavelet transforms, respectively. These features are then fused into a dual Fourier branch, where 1D and 2D Fourier convolutions are applied to the temporal and spatial dimensions, respectively. This, combined with the wavelet module, specifically learns multiple frequency-based features from both temporal and spatial dimensions.}
  
  \label{fig:architecture}
\end{figure*}

 \subsection{Overall Architecture}

As illustrated in Fig. \ref{fig:architecture}, the architecture consists of three key components:
(i) a Temporal Wavelet Convolution module for field frequency disentanglement,
(ii) a Spatial Wavelet Convolution module for spatial frequency decomposition, and
(iii) a Fourier branch block for long-range dependency modeling across decomposed frequency components. Specifically, given an input field $\mathbf{X} \in \mathbb{R}^{T \times C  \times H \times W}$, representing $T$ historical states of multi-field observations, where $H$ is the spatial height and $W$ is the spatial width. The Temporal Wavelet Convolution and Spatial Wavelet Convolution modules are applied to explicitly separate temporal and spatial frequency components. The resulting representations are processed by the Fourier block to predict the next-step field.

\subsection{Spatial Wavelet Convolution}

The SWC module is designed to alleviate spatial spectral bias in neural operators by explicitly decomposing spatial representations into multiple frequency subbands. Unlike conventional convolutions that operate on entangled spatial frequencies, SWC separates global low-frequency structures from localized high-frequency patterns while preserving spatial locality.

We first apply a pointwise convolution to adjust channel dimensions. A 2-D wavelet transform is applied to the input field $\mathbf{X}$. This results in four distinct subband components $\mathbf{X}_{LL}$, $\mathbf{X}_{LH}$, $\mathbf{X}_{HL}$, and $\mathbf{X}_{HH}$. The 2-D wavelet transform is implemented by applying four specific convolutional kernels to each channel of the input with a stride of 1. This process decomposes the image into one low-frequency component ($\mathbf{X}_{LL}$) and three high-frequency components ($\mathbf{X}_{LH}, \mathbf{X}_{HL}, \mathbf{X}_{HH}$), which correspond to the approximation information, horizontal details, vertical details, and diagonal details, respectively. The kernels are defined as follows:

\begin{equation}
    f_{LL} = \frac{1}{9} \begin{bmatrix} 1 & 1 & 1\\ 1 & 1 & 1\\ 1 & 1 & 1\end{bmatrix}, \ 
    f_{LH} = \frac{1}{3} \begin{bmatrix} -1 & 0 & 1 \\ -1 & 0 & 1 \\ -1 & 0 & 1\end{bmatrix},
    \label{f1}
\end{equation}

\begin{equation}
    f_{HL} = \frac{1}{3} \begin{bmatrix} -1 & -1 & -1\\ 0 & 0 & 0 \\ 1 & 1 & 1\end{bmatrix}, \ 
    f_{HH} = \frac{1}{3} \begin{bmatrix} -1 & 0 & 1\\ 0 & 0 & 0\\1 & 0 & -1\end{bmatrix}.
    \label{f2}
\end{equation}
The decomposition process can be formulated as a multi-kernel convolution operation:
\begin{equation}
    \begin{bmatrix} \mathbf{X}_{LL} \\ \mathbf{X}_{LH} \\ \mathbf{X}_{HL} \\ \mathbf{X}_{HH} \end{bmatrix} = \text{Conv} \left( \begin{bmatrix} f_{LL} \\ f_{LH} \\ f_{HL} \\ f_{HH} \end{bmatrix}, \mathbf{X} \right),
\end{equation}
where the convolution is performed with a stride of 1 to ensure that the H/W ratio remains unchanged after convolution and to avoid loss of edge features.

Each of the decomposed frequency subbands is then processed by a convolution layer, utilizing a convolutional kernel $\mathbf{W}_1$ with a size of $3 \times 3$. This specific kernel size is chosen to effectively extract local refined features—such as the front of a shock wave—while preserving the overall spatial resolution. By following this, the feature maps are restored to their initial spatial dimensions through a 2-D inverse wavelet transform (IWT):
\begin{equation}
    \mathbf{X}_{2dwt} = \text{IWT}_{2d} \left( \text{Conv} \left( \mathbf{W}_1, [\mathbf{X}_{LL}, \mathbf{X}_{LH}, \mathbf{X}_{HL}, \mathbf{X}_{HH}] \right) \right).
\end{equation}

The 2-D IWT serves as the reconstruction operator that maps the processed multiresolution features back to the spatial domain. For the Haar wavelet, the IWT essentially reverses the decomposition by interleaving the coefficients. Specifically, for a given set of coefficients at position $(i, j)$ in the subbands, the four corresponding pixels in a $2 \times 2$ block of the reconstructed map are calculated as follows:
\begin{equation}
\begin{cases} 
\begin{aligned}
\mathbf{x}_{2i, 2j} &= \tfrac{1}{2} (\mathbf{X}'_{LL} + \mathbf{X}'_{LH} + \mathbf{X}'_{HL} + \mathbf{X}'_{HH}) \\
\mathbf{x}_{2i+1, 2j} &= \tfrac{1}{2} (\mathbf{X}'_{LL} - \mathbf{X}'_{LH} + \mathbf{X}'_{HL} - \mathbf{X}'_{HH}) \\
\mathbf{x}_{2i, 2j+1} &= \tfrac{1}{2} (\mathbf{X}'_{LL} + \mathbf{X}'_{LH} - \mathbf{X}'_{HL} - \mathbf{X}'_{HH}) \\
\mathbf{x}_{2i+1, 2j+1} &= \tfrac{1}{2} (\mathbf{X}'_{LL} - \mathbf{X}'_{LH} - \mathbf{X}'_{HL} + \mathbf{X}'_{HH})
\end{aligned}
\end{cases}
\end{equation}

In implementation, this is efficiently executed as a transposed convolution with a stride of 2. This reconstruction is vital as it reintegrates localized sharp gradients with the global trend to form a high-fidelity prediction $\mathbf{u}_{t+1}$ without spurious oscillations. In addition, the output of the SWC module employs a customized residual connection:
\begin{equation}
    \mathbf{X}_{Conv2d} = \text{Conv}(\mathbf{W}_2, \text{PWConv}(\mathbf{X}_{input})), 
\end{equation}
\begin{equation}
    \mathbf{X}_{swc} = \mathbf{X}_{2dwt} + \mathbf{X}_{Conv2d}.
\end{equation}
The choice of $3 \times 3$ kernels results in a reduced parameter count, which significantly mitigates the risk of overfitting during the training phase, especially when dealing with limited physical simulation data.

\subsection{Temporal Wavelet Convolution}

The TWC module disentangles slow-varying evolution trends from fast temporal fluctuations to mitigate error accumulation during rollout. Given input field $\mathbf{X}$, we apply a 1-D Haar wavelet transform along the temporal axis, decomposing the feature into a low-frequency component $\mathbf{X}_{L}$ and a high-frequency component $\mathbf{X}_{H}$. 
1D Haar wavelet low-frequency/high-frequency kernels are defined as:
\begin{equation}
    f_L = \frac{1}{k} \cdot [1, 1, \dots, 1] \in \mathbb{R}^{1 \times 1 \times k},
    \label{F4}
\end{equation}
\begin{equation}
    f_H = \frac{1}{k} \cdot [-1, 1, -1, \dots, (-1)^k] \in \mathbb{R}^{1 \times 1 \times k}.
    \label{F3}
\end{equation}
The decomposition process can be formulated as a multi-kernel convolution operation:
\begin{equation}
    \begin{bmatrix} \mathbf{X}_{L} \\ \mathbf{X}_{H} \end{bmatrix} = \text{Conv} \left( \begin{bmatrix} f_{L} \\ f_{H} \end{bmatrix}, \mathbf{X} \right)
\end{equation}
Following the 1-D transform and a pointwise convolution, convolution operations are performed using a kernel $\mathbf{W}_1 \in \mathbb{R}^{1 \times 1 \times 3}$. This configuration diminishes the computational load while enhancing efficiency without compromising simulation fidelity. Finally, the components are reintegrated via a 1-D IWT:
\begin{equation}
    \mathbf{X}_{1dwt} = \text{IWT}_{1d} \left( \text{Conv} \left( \mathbf{W}_1, [\mathbf{X}_{L}, \mathbf{X}_{H}] \right) \right).
\end{equation}

For the Haar basis, the reconstruction of elements $x_{2i}$ and $x_{2i+1}$ from coefficients $X'_{L}$ and $X'_{H}$ is given by:
\begin{equation}
\begin{cases} 
\mathbf{x}_{2i} = \frac{1}{\sqrt{2}} (\mathbf{X}'_{L}(i) + \mathbf{X}'_{H}(i)) \\
\mathbf{x}_{2i+1} = \frac{1}{\sqrt{2}} (\mathbf{X}'_{L}(i) - \mathbf{X}'_{H}(i)) 
\end{cases}
\end{equation}

By reintegrating the low-frequency trend and high-frequency fluctuations, the 1-D IWT ensures that the model maintains field continuity and spectral fidelity. Concurrently, the original feature maps undergo convolution through a layer with kernel $\mathbf{W}_2 \in \mathbb{R}^{1 \times 1 \times 3}$, focusing on local information and relationships between adjacent bands:
\begin{equation}
    \mathbf{X}_{Conv1d} = \text{Conv}(\mathbf{W}_2, \text{PWConv}(\mathbf{X}_{input})).
\end{equation}
The output of the TWC module employs a residual fusion:
\begin{equation}
    \mathbf{X}_{fwc} = \mathbf{X}_{1dwt} + \mathbf{X}_{Conv1d}.
\end{equation}
TWC separately models fast temporal fluctuations and slow-varying trends, capturing temporal evolution more effectively while mitigating error accumulation.

\subsection{Fourier Branch Block}

After explicit spatio-temporal disentanglement via wavelet decomposition, we model long-range dependencies along both temporal and spatial dimensions without re-entangling heterogeneous components. Particularly, we introduce the Fourier branch block, which captures global dependencies while preserving the disentangled structure produced by the wavelet modules.

The block first concatenates the features from two wavelet feature extraction blocks, SWC and TWC, producing
\begin{equation}
    \mathbf{X}_{\text{proj}} = \text{Norm}([\mathbf{X}_{\text{swc}}, \mathbf{X}_{\text{twc}}]),
\end{equation}
The Norm layer normalizes the high-frequency and low-frequency feature distributions. Given the input feature tensor $\mathbf{X}_{\text{proj}}$, the block processes features through two parallel branches: a temporal branch and a spatial branch. Each branch applies Fourier-based convolution along a specific axis, allowing complementary enhancement of temporal and spatial dependencies while respecting the disentangled representation. 
This feature map is subsequently input into the proposed Fourier branch block, i.e., the temporal branch and spatial branch.

\textbf{Temporal Branch.} The temporal branch models dependencies along the time dimension using a one-dimensional Fourier convolution. The input feature $\mathbf{X}_{\text{proj}} \in \mathbb{R}^{T \times C \times H \times W}$ is permuted to $\mathbb{R}^{C \times H \times W \times T}$ and then flattened to $\mathbf{X}_{\text{f}_0} \in \mathbb{R}^{C \cdot H \cdot W \times 1 \times T}$, aligning the temporal dimension with the Fourier transform. The Fourier convolution output is combined with a $1 \times 1$ convolution, followed by GELU activation and residual addition: 
\begin{equation}
    \mathbf{X}_{f_i} = \text{GELU}(\text{SpectralConv}(\mathbf{X}_{f_{i-1}}) + \text{Conv}(\mathbf{X}_{f_{i-1}})) + \mathbf{X}_{f_{i-1}}.
\end{equation}
After several Fourier layers, the output is then reconstructed back to a two-dimensional feature map $\mathbf{X}_{\text{f}} \in \mathbb{R}^{T \times C \times H \times W}$. This design explicitly captures long-range temporal correlations while preserving high-frequency temporal components that are most susceptible to error accumulation, and simultaneously propagates low-frequency trends.

\textbf{Spatial Branch.} Given features $\mathbf{X}_{\mathbf{s}_{0}} = \mathbf{X}_\mathbf{proj}$, the spatial branch captures global dependencies across the spatial H/W dimensions using a two-dimensional Fourier convolution, combined with $1 \times 1$ convolutions and residual connections:
\begin{equation}
    \mathbf{X}_{s_{i}} = \text{GELU}(\text{SpectralConv}(\mathbf{X}_{s_{i-1}}) + \text{Conv}(\mathbf{X}_{s_{i-1}})) + \mathbf{X}_{s_{i-1}}.
\end{equation}
After several Fourier layers, the spatial branch output is $\mathbf{X}_{\text{s}} \in \mathbb{R}^{T \times C \times H \times W}$.
This allows the model to capture spatially long-range interactions while preserving localized high-frequency structures from the wavelet decomposition, alleviating spectral bias.

\textbf{Cross-Branch Fusion.} To integrate temporal and spatial information, the outputs $\mathbf{X}_{f}$ and $\mathbf{X}_{s}$ are concatenated along the channel axis:
$\mathbf{X}_{\text{fuse}} \in \mathbb{R}^{T \times 2C\times H \times W}$. The concatenated features are compressed back to $C$ channels via a $1 \times 1$ convolution, batch normalization, and GELU activation:
\begin{equation}
    \mathbf{X}_{\text{fusion}} = \text{GELU}(\text{BatchNorm}(\text{Conv}(\mathbf{X}_{\text{fuse}}, 1 \times 1))).
\end{equation}
Finally, the fused features are mapped to the desired output channels using another $1 \times 1$ convolution:
\begin{equation}
    \mathbf{X}_{\text{out}} = \text{Conv}(\mathbf{X}_{\text{fusion}}, 1 \times 1) \in \mathbb{R}^{T \times C \times H \times W}.
\end{equation}

By explicitly disentangling spatial and temporal components and applying Fourier convolutions along the respective axes, the Fourier branch block captures long-range dependencies in both space and time. The residual connections, multi-scale wavelet features, and cross-branch fusion ensure that high-frequency local patterns and low-frequency global trends are preserved, enabling accurate and stable modeling of complex, multi-scale spatio-temporal dynamics while maintaining a lightweight network structure.

\begin{table*}[t]\scriptsize
\renewcommand{\arraystretch}{0.92}
\centering
\resizebox{0.99\linewidth}{!}{
\begin{tabular}{llcccccccccccccc}
\toprule
\multirow{2}{*}{\textbf{Model}} & \multirow{2}{*}{\textbf{Metric}} 
& \multicolumn{3}{c}{\textbf{Subcooled}} 
& \multicolumn{3}{c}{\textbf{Gravity}} 
& \multicolumn{3}{c}{\textbf{Saturated}} 
& \multicolumn{3}{c}{\textbf{Single Bubble}} 
& \multicolumn{2}{c}{\textbf{Drug Delivery}}\\
\cmidrule(lr){3-5} \cmidrule(lr){6-8} \cmidrule(lr){9-11} \cmidrule(lr){12-14} \cmidrule(lr){15-16}
& & $\mathbf{T}$ & $\mathbf{V_x}$ & $\mathbf{V_y}$ 
& $\mathbf{T}$ & $\mathbf{V_x}$ & $\mathbf{V_y}$ 
& $\mathbf{T}$ & $\mathbf{V_x}$ & $\mathbf{V_y}$ 
& $\mathbf{T}$ & $\mathbf{V_x}$ & $\mathbf{V_y}$ & $\mathbf{C}$ & $\mathbf{P}$\\
\hline

\rowcolor{blockgray}
\multirow{4}{*}{MIONet} 
& RMSE & 0.036 & 0.137 & 0.097 & 0.064 & 0.394 & \textbf{0.140} & 0.038 & 0.220 & 0.191 & 0.016 & 0.071 & 0.070 & 0.047 & 0.030\\
& FRMSE low & 0.105 & 0.414 & 0.406 & \textbf{0.336} & \textbf{4.108} & \textbf{0.827} & \textbf{0.124} & 0.968 & \textbf{1.193} & 0.169 & 0.843 & 1.131 & -- & --\\
& FRMSE mid & 0.220 & 0.630 & 0.455 & 0.476 & 2.977 & \textbf{0.972} & 0.227 & 1.595 & 1.315 & 0.144 & 0.651 & 0.512 & -- & --\\
& FRMSE high & 0.034 & 0.134 & 0.093 & 0.060 & 0.339 & 0.131 & 0.036 & 0.207 & 0.178 & 0.014 & 0.062 & 0.060 & -- & --\\

\rowcolor{blockgray}
\multirow{4}{*}{U-Net} 
& RMSE & 0.078 & 0.432 & 0.577 & 0.077 & 0.474 & 0.158 & 0.042 & 0.262 & 0.244 & 0.047 & 0.074 & 0.090 & 0.089 & 0.135\\
& FRMSE low & 0.751 & 4.392 & 6.926 & 0.696 & 5.277 & 1.089 & 0.128 & 2.316 & 2.364 & 0.590 & 0.967 & 1.507 & -- & --\\
& FRMSE mid & 0.372 & 1.373 & 1.600 & 0.603 & 3.858 & 1.210 & 0.277 & 1.656 & 1.732 & 0.380 & 0.698 & 0.600 & -- & --\\
& FRMSE high & 0.067 & 0.378 & 0.479 & 0.069 & 0.397 & 0.146 & 0.040 & 0.237 & 0.215 & 0.043 & 0.065 & 0.078 & -- & --\\

\rowcolor{blockgray}
\multirow{4}{*}{CODA-NO} 
& RMSE & 0.047 & 0.149 & 0.116 & 0.076 & 0.635 & 0.148 & 0.048 & 0.416 & 0.425 & -- & -- & -- & 0.147 & 0.076\\
& FRMSE low & 0.235 & 0.712 & 0.836 & 0.609 & 8.358 & 1.288 & 0.332 & 4.960 & 4.730 & -- & -- & -- & -- & --\\
& FRMSE mid & 0.303 & 0.685 & 0.500 & 0.578 & 5.918 & 1.017 & 0.312 & 2.431 & 2.450 & -- & -- & -- & -- & --\\
& FRMSE high & 0.043 & 0.142 & 0.106 & 0.069 & 0.494 & 0.135 & 0.045 & 0.361 & 0.372 & -- & -- & -- & -- & --\\

\rowcolor{blockgray}
\multirow{4}{*}{GNOT} 
& RMSE & 0.047 & 0.260 & 0.244 & 0.118 & 0.538 & 0.428 & 0.048 & 0.435 & 0.474 & \textbf{0.011} & 0.092 & 0.090 & 0.082 & 0.048\\
& FRMSE low & 0.265 & 1.931 & 1.885 & 0.572 & 4.904 & 2.398 & 0.302 & 3.184 & 4.171 & \textbf{0.043} & \textbf{0.182} & 0.305 & -- & --\\
& FRMSE mid & 0.268 & 1.355 & 1.334 & 0.641 & 3.406 & 2.663 & 0.343 & 3.265 & 3.685 & \textbf{0.042} & 0.304 & 0.343 & -- & --\\
& FRMSE high & 0.044 & 0.235 & 0.215 & 0.113 & 0.489 & 0.408 & 0.044 & 0.396 & 0.420 & \textbf{0.009} & 0.094 & 0.080 & -- & --\\

\rowcolor{blockgray}
\multirow{4}{*}{Transolver} & RMSE& 0.083 & 0.139 & 0.099 & 0.082 & 0.651 & 0.147 & 0.099 & 0.342 & 0.402 & 0.085 & 0.175 & 0.059 & 0.236 & 0.091\\
& FRMSE low& 1.278 & 0.517 & 0.422 & 0.701 & 12.572 & 1.272 & 2.384 & 3.029 & 4.281 & 2.519 & 4.174 & 1.808 & -- & --\\
& FRMSE mid& 0.332 & 0.631 & 0.471 & 0.684 & 3.660 & 1.048 & 0.305 & 2.506 & 2.555 & 0.468 & 1.307 & 0.287 & -- & --\\
& FRMSE high& 0.054 & 0.135 & 0.095 & 0.072 & 0.360 & 0.133 & 0.047 & 0.305 & 0.352 & 0.040 & 0.123 & 0.033 & -- & --\\

\rowcolor{blockgray}
\multirow{4}{*}{M2NO} & RMSE& 0.036 & 0.138 & 0.108 & 0.085 & 0.463 & 0.142 & 0.037 & 0.350 & 0.265 & 0.061 & 0.042 & 0.060 & 0.929 & 0.491\\
& FRMSE low& 0.109 & 0.387 & 0.865 & 0.741 & 5.766 & 1.301 & 0.132 & 2.945 & 2.129 & 0.214 & 0.232 & 0.350 & -- & --\\
& FRMSE mid& 0.219 & 0.631 & 0.492 & 0.841 & 3.785 & 0.815 & \textbf{0.218} & 2.605 & 1.747 & 0.226 & 0.157 & 0.231 & -- & --\\
& FRMSE high& 0.034 & 0.134 & 0.095 & 0.071 & 0.380 & 0.131 & 0.036 & 0.313 & 0.243 & 0.057 & 0.031 & 0.040 & -- & --\\

\rowcolor{oursblue}
\multirow{4}{*}{\textbf{WDNO}} 
& RMSE & \textbf{0.028} & \textbf{0.105} & \textbf{0.075} & \textbf{0.060} & \textbf{0.407} & 0.320 & \textbf{0.034} & \textbf{0.154} & \textbf{0.147} & 0.024 & \textbf{0.023} & \textbf{0.026} & \textbf{0.013} & \textbf{0.023}\\
& FRMSE low & \textbf{0.072} & \textbf{0.314} & \textbf{0.339} & 0.438 & 4.802 & 2.851 & 0.409 & \textbf{0.878} & 1.325 & 0.332 & 0.370 & \textbf{0.574} & -- & --\\
& FRMSE mid & \textbf{0.173} & \textbf{0.609} & \textbf{0.319} & \textbf{0.401} & \textbf{2.389} & 1.849 & 0.445 & \textbf{1.401} & \textbf{0.957} & 0.220 & \textbf{0.227} & \textbf{0.219} & -- & --\\
& FRMSE high & \textbf{0.027} & \textbf{0.100} & \textbf{0.072} & \textbf{0.056} & \textbf{0.297} & \textbf{0.116} & \textbf{0.036} & \textbf{0.166} & \textbf{0.137} & 0.012 & \textbf{0.019} & \textbf{0.018} & -- & --\\

\bottomrule
\end{tabular}}
\caption{The Main Results of Multi-Field Rollout Prediction for Multiphase Flow and Drug Delivery Cases.
The CODA-NO model failed to produce effective results for the Single Bubble case due to unstable training. The Drug Delivery case is not suitable for FRMSE testing due to dataset structure issues.
The best results are shown in bold.}
\label{rollout}
\vspace{-0.4cm}
\end{table*}
\section{Experiments}
\textbf{Datasets.} We first adopt four multiphase boiling flow cases \cite{bubbleml}. These datasets exhibit inherent multi-scale temporal dynamics spanning macroscopic phase change to microscale interfacial motion, making them ideal for testing our model’s spatial-temporal disentanglement. They include single bubble boiling, saturated pool boiling, subcooled pool boiling, and gravity flow boiling, with target prediction fields being temperature $\mathbf{T}$ and velocity components $\mathbf{V_x}$, $\mathbf{V_y}$. For generalization validation, we further utilize a respiratory drug delivery dataset \cite{peg}. It follows incompressible Navier–Stokes equations for low-speed viscous airflow, where air flow and drug scalar transport evolve jointly via mass/momentum conservation and advection–diffusion. The predicted quantities here are drug concentration $\mathbf{C}$ and pressure field $\mathbf{P}$.

\textbf{Baseline Models.} We compare WDNO against a comprehensive collection of representative baselines: MIONet \cite{mion}, which builds upon DeepONet; U-Net \cite{unet}, a classic architecture for multiphysics field simulation; transformer-based neural operators including GNOT \cite{GNOT}, CODA-NO \cite{coda-no}, and Transolver \cite{transolver}; as well as M2NO \cite{m2no}, a model that incorporates multi-wavelet and multigrid techniques.

\textbf{Evaluation Metrics.} We employ two metrics to rigorously evaluate prediction accuracy: Root Mean Square Error (RMSE) and Frequency-Dependent Root Mean Square Error (FRMSE)~\cite{bubbleml}.

Multiphysics fields frequently display distinct spatial frequencies, FRMSE is utilized to evaluate WDNO’s ability to capture physical fields across various scales. Specifically, this evaluation metric is defined within the Fourier domain as follows:
$\text{FRMSE} = \sqrt{\frac{\sum^{k_{\max}}_{k = k_{\min}}\left|\mathcal{F}(\hat{f})(k)-\mathcal{F}(f)(k)\right|^2}{k_{\max}-k_{\min}+1}},
$
where $\mathcal{F}$ denotes the discrete Fourier transformation, and $k$ represents the Fourier coordinates. Three wavenumber bands are defined to calculate FRMSE-low, -mid, -high frequency for decoupling multi-scale flow errors and quantifying the model’s performance in capturing cross-scale flow features; detailed settings are provided in the Appendix.

\subsection{Experimental Results and Discussion}

\textbf{Overall Performance.} As shown in Table \ref{rollout}, WDNO achieves state-of-the-art overall performance with the lowest RMSE and FRMSE on nearly all 14 test cases, validating its spatial-temporal disentanglement strategy.
WDNO reduces $\mathbf{T}$ RMSE to 0.028 in the Subcooled case and achieves $\mathbf{V_x}$ and $\mathbf{V_y}$ RMSEs to 0.154 and 0.147, far surpassing CODA-NO in the Saturated case. This proves WDNO balances multi-field optimization weights and prevents dominant variables like temperature from suppressing other physical quantities. For the Drug Delivery case, WDNO attains $\mathbf{C}$ RMSE to 0.013 and $\mathbf{P}$ RMSE to 0.023, while Transolver and M2NO suffer severe error accumulation, owing to WDNO’s advanced spatiotemporal decoupling that suppresses error growth. In the Single Bubble case, WDNO yields $\mathbf{V_x}$ and $\mathbf{V_y}$ RMSEs of 0.023 and 0.026, cutting velocity errors by over 50\% against all baselines. This highlights its strength in resolving steep local flow features sensitive to velocity deviations. Additionally, WDNO is the lightest model tested, striking a favorable accuracy-computation tradeoff.

\textbf{Analysis in Frequency Domain.} WDNO’s capacity to resolve multi-scale features and mitigate field dominance bias is highlighted by the FRMSE results, which are presented in Table \ref{rollout}. WDNO achieved the best performance in most cases.
For lower-energy fields such as $\mathbf{V_x}$ and $\mathbf{V_y}$, the consistent reduction in FRMSE confirms that WDNO’s spatial-temporal disentanglement enhances the learning process for subtle fields, thereby stopping high-energy fields like $\mathbf{T}$ from dominating their patterns. Capturing fine details, the high-frequency regime (FRMSE high) sees WDNO remain competitive, achieving the lowest FRMSE high values in 11 out of all 12 cases. This result indicates that balanced learning enables the retention of high-frequency content while not sacrificing the extraction of geometric features.

\begin{figure*}[!htbp]
  \centering
  \includegraphics[width=1\textwidth]{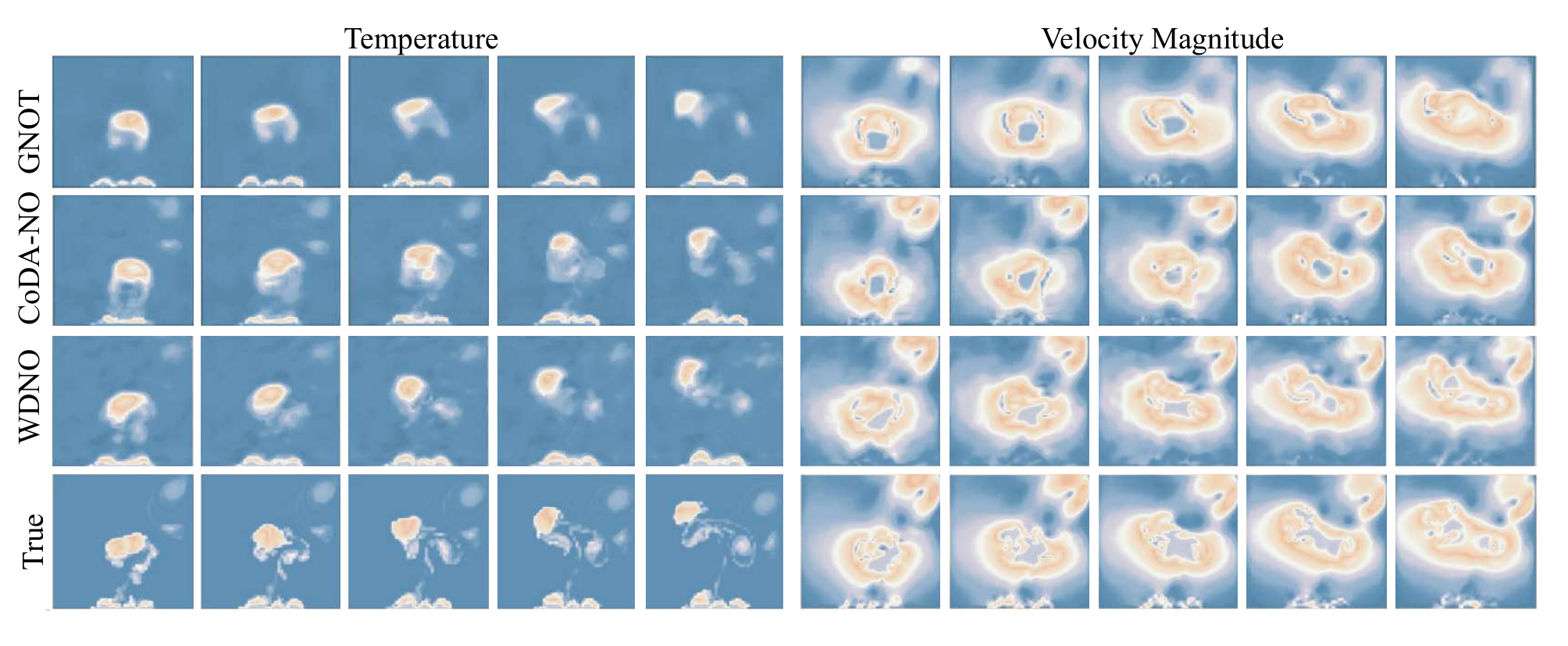}
  \caption{
  The prediction results of GNOT, CoDA-NO and WDNO
  at the 40--45th time steps on the Subcooled case.
  }
  \label{fig:subcooled_rollout}
\end{figure*}

\textbf{Visualization.} As shown in Fig. \ref{fig:subcooled_rollout}, the baseline yields blurry velocity fields with typical field-dominated bias, while WDNO resolves fine flow features including recirculation zones and sharp wall gradients. This confirms that the dual-branch spatial-field feature separation effectively avoids temperature signals from masking velocity dynamics learning. Moreover, the baseline suffers from error accumulation at critical time steps, causing small eddy dissipation and bubble deformation, whereas WDNO maintains excellent spatiotemporal consistency and accurately tracks vapor phase evolution and moving contact line. Visualizations of all baselines are available in the Appendix.

\subsection{Noise Robustness Analysis} We design experiments to test the robustness of WDNO and baseline models by polluting training inputs with Gaussian noise of varying strengths; results are listed in Table \ref{noise}. Experiments verify WDNO’s outstanding robustness. RMSE barely rises from 0.346 (1\% noise) to 0.339 (5\% noise), retaining the lowest error throughout. All baseline models degrade visibly under stronger noise: MIONet’s RMSE climbs from 0.478 to 0.505, while Transolver’s jumps from 0.401 to 0.608. Such stable superior performance under noisy training data stems from WDNO’s strong universal learning capacity for physical fields.

\begin{table}[t]
\renewcommand{\arraystretch}{1}
\centering
\resizebox{1\linewidth}{!}{
\begin{tabular}{c c c c c c c c}
\hline
 & MIONet & U-Net & CODA-NO & GNOT & Transolver & M2NO & WDNO \\
Noise level 
& RMSE 
& RMSE 
& RMSE 
& RMSE 
& RMSE 
& RMSE
& RMSE\\
\hline
\rowcolor{blockgray}
1\% & 0.478 & 0.375 & 0.412 & 0.432 & 0.401 & 0.335 &\textbf{0.346} \\
3\% & 0.491 & 0.382 & 0.421 & 0.441 & 0.481 & 0.356 &\textbf{0.341} \\
\rowcolor{blockgray}
5\% & 0.505 & 0.395 & 0.423 & 0.445 & 0.608 & 0.392 &\textbf{0.339} \\
\hline
\end{tabular}
}
\caption{The main results of multi-field prediction for Subcooled multiphase flow cases with noisy data}
\label{noise}
\end{table}

\begin{table}[h]
\renewcommand{\arraystretch}{1}
\begin{center}
\resizebox{1\linewidth}{!}{
\begin{tabular}{ccccccccc}
\hline
 & & MIONet & U-Net & CODA-NO & GNOT & Transolver & M2NO & WDNO \\
Fields & Resolution & RMSE & RMSE & RMSE & RMSE & RMSE & RMSE & RMSE \\
\hline
\rowcolor{blockgray}
     &1x&  0.048 &  0.036 &  0.037 &  0.038 & 0.038 & 0.034 & \textbf{0.033} \\
$\mathbf{T}$        &2x&  - &      0.057 &  0.040 &  0.038 & - & 0.037 & \textbf{0.033} \\
        &4x&  - &      0.060 &  0.059 &  0.038 & - & 0.048 & \textbf{0.032} \\
\rowcolor{blockgray}
   &1x&  0.231 &  0.185 &  0.201 &  0.211 & 0.188 & 0.162 & \textbf{0.171} \\
$\mathbf{V_x}$        &2x&  - &      0.213 &  0.195 &  0.199 & - & 0.196 & \textbf{0.169} \\
        &4x&  - &      0.228 &  0.215 &  0.194 & - & 0.228 & \textbf{0.169} \\
\rowcolor{blockgray}
   &1x&  0.221 &  0.165 &  0.178 &  0.217 & 0.162 & 0.136 & \textbf{0.142} \\
$\mathbf{V_y}$        &2x&  - &      0.226 &  0.193 &  0.215 & - & 0.180 & \textbf{0.139} \\
        &4x&  - &      0.267 &  0.238 &  0.217 & - & 0.215 & \textbf{0.138} \\
\hline
\end{tabular}
}
\end{center}
\caption{The results of super-resolution evaluation on the Subcooled case. MIONet and Transolver do not support super-resolution due to their architecture.}
\label{super}
\end{table}

\begin{table*}[!htbp]
\renewcommand{\arraystretch}{1}
\centering
\resizebox{1\linewidth}{!}{
\begin{tabular}{llcccccccccccccc}
\hline
\multirow{2}{*}{\textbf{Model}} & \multirow{2}{*}{\textbf{Metric}} 
& \multicolumn{3}{c}{\textbf{Subcooled}} 
& \multicolumn{3}{c}{\textbf{Gravity}} 
& \multicolumn{3}{c}{\textbf{Saturated}} 
& \multicolumn{3}{c}{\textbf{Single Bubble}} 
& \multicolumn{2}{c}{\textbf{Drug Delivery}}\\
\cline{3-16}
& & $\mathbf{T}$ & $\mathbf{V_x}$ & $\mathbf{V_y}$ 
& $\mathbf{T}$ & $\mathbf{V_x}$ & $\mathbf{V_y}$ 
& $\mathbf{T}$ & $\mathbf{V_x}$ & $\mathbf{V_y}$ 
& $\mathbf{T}$ & $\mathbf{V_x}$ & $\mathbf{V_y}$ 
& $\mathbf{C}$ & $\mathbf{P}$\\
\hline

\rowcolor{blockgray}
\multirow{4}{*}{FNO}
& RMSE & 0.036 & 0.187 & 0.167 & 0.121 & 0.551 & 0.436 & 0.052 & 0.452 & 0.468 & 0.014 & 0.084 & 0.098 & 0.027 & 0.027 \\
& FRMSE low & 0.140 & 0.921 & 0.960 & 0.548 & 4.019 & 1.671 & 0.265 & 3.486 & 3.668 & 0.044 & 0.217 & 0.335 & -- & --\\
& FRMSE mid & 0.200 & 0.984 & 0.940 & 0.679 & 3.398 & 2.831 & 0.376 & 3.559 & 3.734 & 0.055 & 0.265 & 0.376 & -- & --\\
& FRMSE high & 0.034 & 0.175 & 0.151 & 0.113 & 0.509 & 0.422 & 0.048 & 0.403 & 0.415 & 0.014 & 0.074 & 0.077 & -- & --\\

\rowcolor{blockgray}
\multirow{4}{*}{w/o TWC}
& RMSE & 0.034 & 0.170 & 0.147 & 0.126 & 0.498 & 0.424 & 0.049 & 0.391 & 0.398 & 0.007 & 0.033 & 0.040 & 0.006 & 0.010 \\
& FRMSE low & 0.136 & \textbf{0.674} & 0.724 & 0.568 & 3.039 & 1.262 & 0.224 & 2.429 & 2.644 & 0.091 & 0.268 & 0.594 & -- & --\\
& FRMSE mid & 0.193 & 0.909 & 0.859 & 0.667 & 3.095 & 2.633 & 0.366 & 3.231 & 3.347 & 0.053 & 0.279 & 0.327 & -- & --\\
& FRMSE high & 0.032 & 0.161 & 0.136 & 0.122 & 0.469 & 0.409 & 0.046 & 0.355 & 0.358 & 0.006 & 0.031 & 0.035 & -- & --\\

\rowcolor{blockgray}
\multirow{4}{*}{w/o SWC}
& RMSE & 0.033 & 0.186 & 0.142 & 0.117 & 0.509 & 0.422 & 0.049 & 0.372 & 0.388 & 0.007 & 0.056 & 0.037 & 0.005 & 0.018\\
& FRMSE low & 0.122 & 0.886 & 0.702 & 0.477 & 3.253 & 1.194 & 0.191 & 2.226 & 2.483 & 0.100 & 0.689 & 0.532 & -- & --\\
& FRMSE mid & 0.188 & 0.983 & 0.826 & 0.667 & 3.169 & 2.657 & 0.351 & 3.038 & 3.243 & 0.047 & 0.516 & 0.304 & -- & --\\
& FRMSE high & 0.032 & 0.174 & 0.132 & 0.113 & 0.477 & 0.408 & 0.047 & 0.339 & 0.351 & 0.006 & 0.049 & 0.032 & -- & --\\

\rowcolor{blockgray}
\multirow{4}{*}{w/o DualBranch}
& RMSE & 0.034 & 0.184 & 0.162 & 0.118 & 0.495 & 0.416 & 0.049 & 0.391 & 0.396 & 0.007 & 0.033 & 0.037 & 0.005 & 0.008 \\
& FRMSE low & 0.126 & 0.809 & 0.862 & 0.517 & 3.033 & 1.138 & 0.198 & 2.432 & 2.539 & 0.054 & 0.237 & 0.447 & -- & --\\
& FRMSE mid & 0.195 & 0.975 & 0.929 & 0.643 & 3.116 & 2.512 & 0.364 & 3.218 & 3.341 & 0.051 & 0.277 & 0.316 & -- & --\\
& FRMSE high & 0.033 & 0.173 & 0.149 & 0.111 & 0.465 & 0.403 & 0.046 & 0.355 & 0.356 & 0.006 & 0.031 & 0.033 & -- & --\\

\rowcolor{oursblue}
\multirow{4}{*}{\textbf{WDNO}}
& RMSE
& \textbf{0.033} & \textbf{0.171} & \textbf{0.142}
& \textbf{0.041} & \textbf{0.184} & \textbf{0.164}
& \textbf{0.046} & \textbf{0.361} & \textbf{0.361}
& \textbf{0.005} & \textbf{0.020} & \textbf{0.019} & \textbf{0.003} & \textbf{0.005}\\

& FRMSE low
& \textbf{0.119} & 0.763 & \textbf{0.694}
& \textbf{0.207} & \textbf{1.056} & \textbf{1.122}
& \textbf{0.223} & \textbf{2.217} & \textbf{2.463}
& \textbf{0.032} & \textbf{0.183} & \textbf{0.256} & -- & --\\

& FRMSE mid
& \textbf{0.180} & \textbf{0.857} & \textbf{0.773}
& \textbf{0.289} & \textbf{1.258} & \textbf{1.279}
& \textbf{0.321} & \textbf{2.843} & \textbf{2.980}
& \textbf{0.030} & \textbf{0.132} & \textbf{0.139} & -- & --\\

& FRMSE high
& \textbf{0.031} & \textbf{0.159} & \textbf{0.128}
& \textbf{0.040} & \textbf{0.175} & \textbf{0.151}
& \textbf{0.044} & \textbf{0.330} & \textbf{0.333}
& \textbf{0.004} & \textbf{0.017} & \textbf{0.016} & -- & --\\

\hline
\end{tabular}}
\caption{The RMSE and FRMSE of multi-field prediction across four coupled multiphysics cases. The models evaluated included FNO, WDNO without TWC module, WDNO without SWC module, and WDNO without dual branch structure.}
\label{xr}
\end{table*}

\subsection{Super-Resolution Performance} We assess model generalization for super-resolution on high-resolution test set, and the evaluation results are shown in Table \ref{super}. WDNO yields nearly constant RMSE across 1x/2x/4x: $\mathbf{T}$ error hovers 0.032–0.033, with $\mathbf{V_x}$ and $\mathbf{V_y}$ RMSEs limited to 0.169–0.171 and 0.138–0.142. Conversely, U-Net, CODA-NO and M2NO exhibit sharp error surges at higher resolutions. These results validate WDNO’s outstanding resolution robustness. Its integrated spatial-temporal Haar wavelet modules separate multi-frequency components, preserving low-frequency global patterns and recovering high-frequency fine textures. The frequency-decoupled framework maintains stable accuracy under arbitrary output resolutions, proving the superiority of wavelet-based spatial-temporal disentanglement in multi-scale dynamic field prediction.

\begin{figure}[!ht]
\begin{center}
\includegraphics[width=0.49\textwidth]{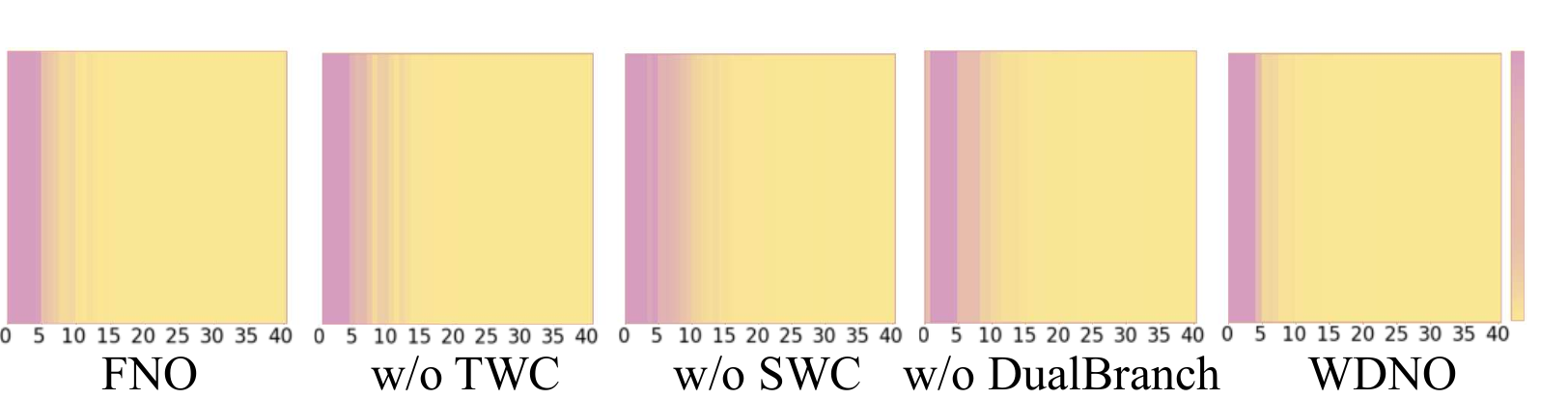}
\caption{The cumulative spectral energy errors of all fields across different wave number modes. The upper part of the color bar represents a larger error.}\label{fig:spectral_energy}
\vspace{-0.6cm}
\end{center}
\end{figure}

\subsection{Ablation Study}
Table \ref{xr} shows the module ablation results, each module plays an irreplaceable role. As observed, w/o TWC, SWC, and DualBranch, the performance across the 12 metrics varies, and all are superior to the base FNO. Removing the TWC module severely degrades temporal modeling, increasing the RMSE of $\mathbf{T}$ from 0.041 to 0.126 in the Gravity case. Eliminating the SWC module impairs spatial reconstruction, raising the RMSE of $\mathbf{P}$ from 0.005 to 0.018 in the Drug Delivery case. Disabling the dual-branch structure weakens spatiotemporal fusion, increasing the RMSE of $\mathbf{V_y}$ from 0.142 to 0.162 in the Subcooled case. The consistent performance decline of all variants demonstrates the rationality of the WDNO design.

Fig. \ref{fig:spectral_energy} further indicates that each module reduces cumulative spectral energy error and performs complementary, non-overlapping spectral optimization across different wavenumber domains. Each module targets unique spectral features and compensates for the deficiencies of other branches, confirming their independent and distinctive roles in feature learning.

\subsection{Computational Cost} 

To evaluate the computational efficiency of the proposed WDNO model, we quantitatively compare it against representative baselines in Table \ref{fl}, which reports single-forward inference latency (seconds) and floating-point operations (MFLOPs). WDNO achieves the fastest inference speed at only 0.08 s, substantially outperforming MIONet (1.21 s), M2NO (1.25 s), CODA-NO (1.52 s), and GNOT (0.30 s). It also achieves the lowest computational complexity of merely 38 MFLOPs, which is compared with all baselines including CODA-NO (4855 M), GNOT (2593 M), M2NO (10966 M), and u-Net (484 M), yielding 77\%–99\% lower FLOPs. Powered by frozen wavelet kernels for feature extraction, WDNO surpasses even FNO in efficiency. Overall, WDNO dominates all compared methods in both inference speed and computational cost, striking a favorable balance between predictive accuracy and efficiency. These results validate its lightweight design, making WDNO suitable for real-time inference in practical engineering and a viable option for complex multiphysics simulations.

\begin{table}[h]
\renewcommand{\arraystretch}{1}
\begin{center}
\resizebox{1\linewidth}{!}{
\begin{tabular}{ccccccccc}
\hline
 & MIONet & U-Net & CODA-NO & GNOT & Transolver & M2NO& WDNO \\
\hline
\rowcolor{blockgray}
Inference Time & 1.21 & 0.78 & 1.52 & 0.30 & 1.18 & 1.25 &0.08 \\
FLOPs          & 211  & 484  & 4855 & 2593 & 184 & 10966 &38  \\
\hline
\end{tabular}
}
\end{center}
\caption{The inference time per forward propagation and floating-point operations (FLOPs) of all models}
\label{fl}
\end{table}

\subsection{One-step Prediction}

Beyond rollout prediction, we conduct one-step prediction to evaluate short-term forecasting (Table in Appendix). WDNO achieves the best RMSE among all baselines: for Gravity, it reduces \(\mathbf{V}_\mathbf{x}\) and \(\mathbf{V}_\mathbf{y}\) RMSE by nearly 60\% compared with U-Net (0.792 and 0.425). For Drug Delivery, its RMSEs reach 0.003 for \(\mathbf{C}\) and 0.005 for \(\mathbf{P}\), outperforming Transolver (0.120 for \(\mathbf{C}\) and 0.097 for \(\mathbf{P}\)). In FRMSE, WDNO outperforms all baselines across frequencies, especially at high frequencies. For Single Bubble, its high-frequency FRMSEs are 0.005 (\(\mathbf{T}\)), 0.020 (\(\mathbf{V}_\mathbf{x}\)), and 0.019 (\(\mathbf{V}_\mathbf{y}\)), versus M2NO’s 0.050, 0.123, and 0.160. It also improves low-frequency predictions notably, capturing high-frequency features well while retaining strong low-frequency modeling capacity with no obvious trade-off.
\section{Conclusions}
This work presents a neural operator framework with explicit spatial-temporal disentanglement for evolving physical field modeling. It adopts 2D/1D wavelet decomposition for spatial and temporal modeling to preserve fine local features and mitigate spectral bias and long-term prediction error accumulation. A Fourier branch further captures long-range spatio-temporal dependencies via parallel learning and cross-branch fusion. Evaluated on multiphysics benchmarks, the method accurately models multi-scale spatial patterns and temporal dynamics, providing an efficient operator learning solution for scientific computing.

\bibliographystyle{IEEEtran}
\bibliography{main}
\clearpage
\appendix

\section{Related Work}

Neural operators have emerged as a transformative paradigm for approximating mappings between infinite-dimensional function spaces for solving PDEs. They can be broadly categorized into the following types: 

1) Spectral-Domain Operators: The FNO \cite{fno} pioneered learning in the spectral domain through Fourier transforms, enabling efficient parameter-to-solution mappings in frequency space. Subsequent extensions, such as Geo-FNO \cite{li2023fourier}, address complex geometries by incorporating coordinate transformations \cite{gno}, while domain decomposition variants \cite{Zhou2025Rapid, Kossaif2024Multi} facilitate multi-scale resolution. Although highly computationally efficient, these spectral methods face inherent limitations when dealing with discontinuous fields and heterogeneous input modalities. 

2) Branch-Trunk Architectures: DeepONet \cite{deeponet} established a foundational framework that decouples input processing via a branch network from query point encoding via a trunk network, offering universal approximation guarantees. Physics-informed variants \cite{wang2021learning, wang2022improved} enhance generalization capabilities by integrating PDE-constrained optimization, and MIONet \cite{mion} extends this architecture to handle systems with multiple input functions. While demonstrating notable flexibility across various PDEs, these methods typically encounter scaling challenges when applied to high-dimensional parameter spaces. 

3) Transformer-Based Operators: Recent advancements have leveraged transformer architectures to achieve enhanced geometric adaptability in neural operators. OFormer addresses multi-scale challenges through structured attention patterns~\cite{li2023transformer}. Galerkin Transformer \cite{cao2021choose}, for instance, reinterprets attention mechanisms as learnable integral kernels, establishing theoretical connections to traditional operator learning. CoDA-NO \cite{coda-no} introduced a breakthrough by treating physical variables as infinite-dimensional tokens, enabling native handling of function space inputs. More recently, Transolver \cite{transolver} adaptively divides a discrete grid into learnable physical slices and computes physical perception tokens using Physics-Attention.

4) Wavelet neural operator: Wavelet Neural Operator (WNO) \cite{xb2} is a family of neural operators built upon the wavelet transform. It addresses the limitation of Fourier Neural Operators in spatial-frequency localization. Equipped with superior time-frequency localization and multi-scale analysis abilities, WNO and its variants excel at learning mappings across function spaces, and achieve promising performance in solving partial differential equations (PDEs) with local abrupt changes, processing non-stationary signals, and modeling multi-scale physical fields \cite{xb1,xb3,xb4}. The recently proposed M2NO \cite{m2no} integrates diverse wavelet transforms and multi-grid structures, further boosts the overall solving accuracy and efficiency.

\section{Problem Formulation}

We consider the problem of learning the evolution dynamics of multiple coupled physical fields governed by PDEs. Let $\Omega \subset \mathbb{R}^d$ denote the spatial domain and $t \in [0,T]$ the temporal interval. At each time step, the system state is represented by a collection of physical fields
$
\mathbf{u}(\cdot,t) = \{u^{(1)}(\cdot,t), \ldots, u^{(m)}(\cdot,t)\},
$
where $u^{(i)}: \Omega \rightarrow \mathbb{R}$ corresponds to the $i$-th physical quantity.

Our goal is to learn an evolution operator $\mathcal{G}_\Delta$ such that
\begin{equation}
\mathbf{u}(\cdot,t+\Delta t) = \mathcal{G}_\Delta\big(\mathbf{u}(\cdot,t)\big),
\end{equation}
using observational or simulated data. In practice, we adopt an autoregressive neural operator that predicts the next-step solution field based on a short temporal history,
\begin{equation}
\mathcal{G}_\Delta: \mathbf{u}_{t-k:t} \mapsto \mathbf{u}_{t+1},
\end{equation}
and recursively rolls out the learned operator to obtain long-horizon predictions.

\section{Pseudocode}

\subsubsection{Temporal Wavelet Convolution(TWC)}

Input feature tensor dimension: $X \in \mathbb{R}^{T \times C \times H \times W}$, where:
\begin{itemize}
    \item $T$: Historical
states of multi-field observations, $C$: Number of channels, $H$: Spatial height, $W$: Spatial width;
    \item Wavelet convolution kernel size $k=3$, padding satisfies $\text{pad} = \lfloor k/2 \rfloor = 1$ (ensures input/output size consistency along the temporal dimension).
\end{itemize}
1D Haar wavelet low-frequency/high-frequency kernels are defined as:
\begin{align}
    \mathbf{K}_{\text{low}} &= \frac{1}{k} \cdot [1, 1, \dots, 1] \in \mathbb{R}^{1 \times 1 \times k}, \\
    \mathbf{K}_{\text{high}} &= \frac{1}{k} \cdot [-1, 1, -1, \dots, (-1)^k] \in \mathbb{R}^{1 \times 1 \times k}.
\end{align}
Kernel parameters are frozen during training ($\text{requires\_grad} = \text{False}$).

\begin{algorithm}[!ht]
\caption{TWC Module}
\label{alg:swc_forward}
\textbf{Input}: $X \in \mathbb{R}^{T \times C \times H \times W}$, wavelet kernels $f_L, f_H$\\
\textbf{Output}: $X_{\text{out}} \in \mathbb{R}^{T \times C_{\text{out}} \times H \times W}$
\begin{algorithmic}[1]
        \STATE $X = Reshape(X)$ // $[C \times H \times W, 1, T]$

        \STATE $X_L = Conv_{f_L}{X}$ // Low-frequency component, $[C \times H \times W, 1, T]$
        \STATE $X_H = Conv_{f_H}{X}$ // High-frequency component, $[C \times H \times W, 1, T]$

        \STATE $X_L = Reshape(X_L)$ // $[T, C, H, W]$
        \STATE $X_H = Reshape(X_H)$ // $[T, C, H, W]$

        \STATE $X_{\text{pw}} = Conv_{1 \times 1}{X}$ // $[T, C_{\text{out}}, H, W]$
        \STATE $X_{\text{L\_pw}} = Conv_{1 \times 1}{X_L}$ // Low-frequency component downsampling
        \STATE $X_{\text{H\_pw}} = Conv_{1 \times 1}{X_H}$ // High-frequency component downsampling

        \STATE $X_{\text{res}} = Conv{}{X_{\text{pw}}}$ // $[B, C_{\text{out}}, H, W]$

        \STATE $X_{\text{freq}} = \text{IWT}_{1d}([X_{\text{L\_pw}}, X_{\text{H\_pw}}], \text{dim}=1)$ // {$[B, C_{\text{out}}, H, W]$}

        \STATE $X_{\text{out}} = X_{\text{freq}} + X_{\text{res}}$\\
        \STATE \textbf{return} $X_{\text{out}}$
    \end{algorithmic}
\end{algorithm}

\subsubsection{Spatial Wavelet Convolution (SWC)}

The Spatial Wavelet Convolution (SWC) module performs 2D wavelet decomposition along the spatial dimension ($H \times W$), including four frequency components (low-low, low-high, high-low, high-high). Its core definition is as follows:

2D Haar wavelet kernels ($3 \times 3$) are defined as:
\begin{align}
    \mathbf{K}_{\text{ll}} &= \frac{1}{9} \cdot
    \begin{bmatrix}
        1 & 1 & 1 \\
        1 & 1 & 1 \\
        1 & 1 & 1
    \end{bmatrix}, \quad
    \mathbf{K}_{\text{lh}} = \frac{1}{3} \cdot
    \begin{bmatrix}
        -1 & 0 & 1 \\
        -1 & 0 & 1 \\
        -1 & 0 & 1
    \end{bmatrix}, \\
    \mathbf{K}_{\text{hl}} &= \frac{1}{3} \cdot
    \begin{bmatrix}
        -1 & -1 & -1 \\
        0 & 0 & 0 \\
        1 & 1 & 1
    \end{bmatrix}, \quad
    \mathbf{K}_{\text{hh}} = \frac{1}{3} \cdot
    \begin{bmatrix}
        -1 & 0 & 1 \\
        0 & 0 & 0 \\
        1 & 0 & -1
    \end{bmatrix}.
\end{align}

\begin{algorithm}[!ht]
\caption{SWC Module}
\label{alg:wsc_forward}
\textbf{Input}: $X \in \mathbb{R}^{T \times C \times H \times W}$, 2D wavelet kernels $f_{LL},f_{LH},f_{HL},f_{HH}$\\
\textbf{Output}: $X_{\text{out}} \in \mathbb{R}^{T \times C_{\text{out}} \times H \times W}$
\begin{algorithmic}[1]
\STATE $X_{\text{pw}} = Conv_{1 \times 1}(X)$
\STATE $X_{\text{res}} = Conv{}(X_{\text{pw}})$
\STATE $X_{LL} = Conv_{f_{LL}}(X_{\text{pw}})$
\STATE $X_{LH} = Conv_{f_{LH}}(X_{\text{pw}})$
\STATE $X_{HL} = Conv_{f_{HL}}(X_{\text{pw}})$
\STATE $X_{HH} = Conv_{f_{HH}}(X_{\text{pw}})$
\STATE $X_{\text{freq}} = IWT2(Conv ([X_{\text{LL}}, X_{\text{LH}}, X_{\text{HL}}, X_{\text{HH}}]))$ 
\STATE $X_{\text{out}} = X_{\text{freq}} + X_{\text{res}}$\\
\STATE \textbf{return} $X_{\text{out}}$
\end{algorithmic}
\end{algorithm}

\subsection{Fourier Branch Block}
Before the Fourier Branch Block, we integrate TWC and SWC modules, completing feature fusion via concatenation, Batch Normalization (BN), and GeLU activation. It is defined as:
\begin{align}
    X_{\text{temporal}} &= \text{TWC}(X), \\
    X_{\text{spatial}} &= \text{SWC}(X), \\
    X_{\text{proj}} &= \text{GeLU}(\text{BN}(\text{concat}([X_{\text{spectral}}, X_{\text{spatial}}], \text{dim}=1))).
\end{align}

\begin{algorithm}[!htb]
\caption{Fourier Branch Block}
\textbf{Input}: $X_{proj} \in \mathbb{R}^{T \times C \times H \times W}$\\
\textbf{Parameter}: $C_{\text{width}}$ (hidden channel width), $n_{\text{layers}}$ (number of layers)\\
 \textbf{Output}: $X_{out} \in \mathbb{R}^{T \times C \times H \times W}$
\begin{algorithmic}[1]
        \STATE $X_{temporal} = X_{proj}$
        \STATE $X_{spatial} = X_{proj}$

        \STATE $X_{temporal} = Reshape({{X_{temporal}}})$ // $[CHW,1,T]$ Temporal Branch Shape Transformation

        \FOR{$i = 0, 2, ..., 2n_{\text{layers}}-2$}
            \STATE ${X_{temporal_{ft}}} = {SpectralConv}({X_{temporal}})$ // Temporal Branch (1D FNO)
            \STATE ${X_{temporal_{conv}}} = {Conv}({X_{temporal}})$
            \STATE $X_{temporal} = {GELU}({X_{temporal_{ft}}} + {X_{temporal_{conv}}}) + X_{temporal}$
  
            \STATE ${X_{spatial_{ft}}} = {SpectralConv}({X_{spatial}})$ // Spatial Branch (2D FNO)
            \STATE ${X_{spatial_{conv}}} = {Conv}({X_{spatial}})$
            \STATE $X_{spatial} = {GELU}({X_{spatial_{ft}}} + {X_{spatial_{conv}}}) + X_{spatial}$
        \ENDFOR

        \STATE $X_{temporal} = Reshape({X_{temporal}})$ // $[T \times C_{\text{width}} \times H \times W]$  Temporal Branch Shape Recovery

        \STATE ${X_{fuse}} = {Cat}([X_{temporal}, X_{spatial}], \text{dim}=1)$ // $[T \times 2C_{\text{width}} \times H \times W]$
        \STATE ${X_{fusion}} = {GELU}(BatchNorm(Conv_{1 \times 1}({X_{fuse}})))$

        \STATE $X_{out} = Conv(X_{fusion})$\\
        \STATE \textbf{return} ${X_{out}}$
        
    \end{algorithmic}
\end{algorithm}

\section{Experiment Detail} 

\subsection{Settings} The AdamW optimizer was utilized for the training process of all models, with the training duration set to 50 epochs in four bubble cases and 10 epochs in the Drug Delivery case. When it comes to the parameter settings of the models, all of them allowed for adjustments in network width (32 or 64), layer count (2 or 4), and batch size (2, 4, or 8). The number of attention heads for the transform-based model is set to 4 layers.
Specific parameter configurations differed marginally across the various model types. For the U-Net model, a lightweight encoder was adopted to effectively capture geometric interface information, and this encoder could be configured with either 2 or 4 layers. In contrast, the FNO model featured a fixed network width of 32, and its mode count offered two selectable options: 8 and 16. 

In terms of dataset preprocessing for training, the Subcooled, Gravity, and Saturated cases were each downsampled by a factor of 8. On the other hand, the Single Bubble case was downsampled by a factor of 2 for training purposes. For super-resolution experiments, the dataset was downsampled by factors of 4 and 2 separately, which were then used as the test sets corresponding to the 4x and 2x super-resolution tasks.

A unified rule was applied to split all datasets into training and test subsets. The Single Bubble case, which contained 500 time steps, was divided such that 450 time steps were allocated to the training set and the remaining 50 to the test set. For the Subcooled, Gravity, Saturated and Drug Delivery cases-each of which included multiple simulation trajectories-one trajectory was designated as the test subset for each dataset, while all other trajectories were used for model training. The spatial structure of the Drug Delivery case is mapped to 2D to meet the model's input requirements.

In noise analysis, the noise standard deviation is defined as $\sigma = \nu \cdot (\max(x) - \min(x))$, where $x$ denotes field points and $\nu$ stands for noise level.

\subsection{Learning Objective} The primary objective of the model optimization is to maximize the predictive accuracy with respect to the ground truth data. To this end, the training objective is framed as the minimization of the Mean Squared Error (MSE), which quantifies the L2-norm distance between the model's predictions and the target values. 
\begin{equation}
    \mathcal{L} = \frac{1}{|F|}\sum_{\mathbf{x} \in F}(\hat{f}(\mathbf{x})-f(\mathbf{x}))^2,
\end{equation}
where $F$ denotes all points in the physical field, $\hat{f}(\mathbf{x})$ represents the model's output at location $\mathbf{x}$, and $f(\mathbf{x})$ is the target physical field value.

\subsection{Metrics Detail} \label{a metrics}Across the entire computational domain, the RMSE serves as the key indicator for assessing how accurate the predicted physical field is overall. The specific method for calculating this metric is given by the following formula:
$\text{RMSE} = \sqrt{\frac{1}{|F|}\sum_{\mathbf{x} \in F}(\hat{f}(\mathbf{x})-f(\mathbf{x}))^2}$.

For FRMSE, we define three ranges for $k_{\min}$ and $k_{\max}$ to isolate errors across scales:
\begin{itemize}
    \item {FRMSE (low)}: ${[0, 4]}$, capturing large-scale flow features.
    \item {FRMSE (mid)}: ${[5, 12]}$, capturing medium-scale features.
    \item {FRMSE (high)}: ${[13, \infty]}$, representing fine-grained details and small-scale features.
\end{itemize}
In different spectral domains, the magnitude of errors is quantified by these value ranges, which in turn provide valuable insights into the model’s effectiveness in capturing the detailed characteristics of multi-scale fields.

\subsection{Visualization} 
\begin{figure*}[!htbp]
  \centering
  \includegraphics[width=1\textwidth]{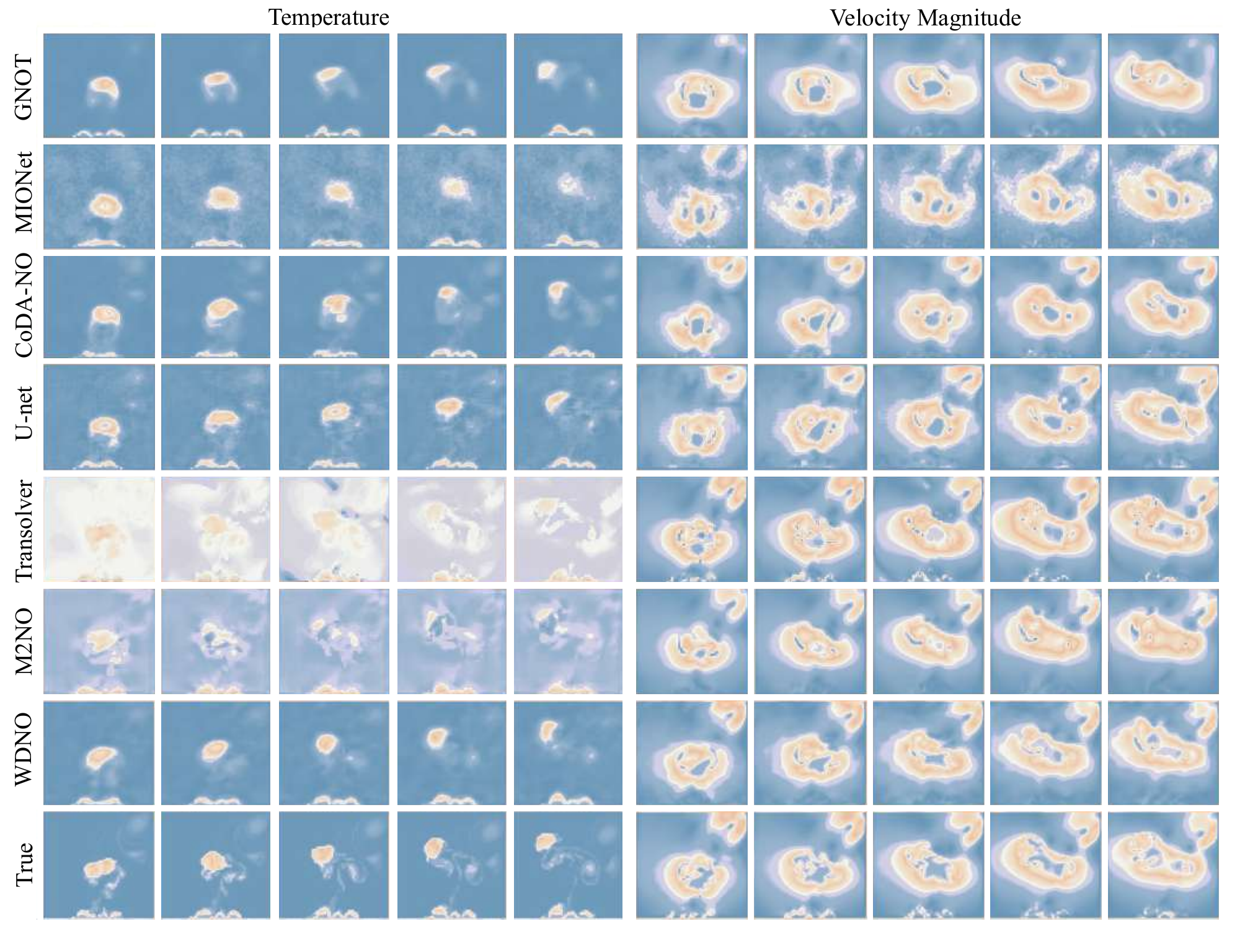}
  \caption{
  The prediction results of U-net, MIONet, CoDA-NO, GNOT, and WDNO
  at the 40--45th time steps on the Subcooled case.
  }
  \label{fig:subcooled_rollout1}
\end{figure*}

As illustrated in Fig. \ref{fig:subcooled_rollout1}, the velocity fields predicted by U-Net, MIONet, CoDA-NO and GNOT exhibit severe error accumulation, whose contours deviate substantially from the ground truth. Although Transolver and M2NO achieve relatively low accumulated errors, they suffer from global bias in temperature field predictions. In contrast, WDNO yields minor cumulative errors with balanced prediction performance across both temperature and velocity fields, delivering the optimal overall forecasting accuracy.

\subsection{Computational Cost} 

To evaluate the computational efficiency of the proposed WDNO model, a quantitative comparison with several representative baselines is conducted in Table \ref{fl}. This table summarize two critical indicators: inference time (in seconds) for a single forward pass and the corresponding floating-point operations (FLOPs, in millions). In terms of inference efficiency, the WDNO model achieves the most favorable inference speed, with a single forward propagation completed in merely 0.08 seconds. This latency is remarkably lower than that of baseline models such as MIONet (1.21 s), M2NO (1.25 s), CODA-NO (1.52 s), and even the relatively efficient GNOT (0.30 s), representing a 95\% reduction compared with CODA-NO. As for computational complexity, the WDNO model only involves 38 MFLOPs, which is reduced by an order of magnitude compared with all other baseline approaches, drastically lower than those of CODA-NO (4855 M), GNOT (2593 M), M2NO (10966 M), and u-Net (484 M), with FLOPs decreased by approximately 77\% to 99\% relative to all benchmarks. 

Benefiting from the frozen wavelet kernel as a feature extraction module, WDNO's computational efficiency even surpasses that of FNO. In summary, the WDNO model exhibits clear superiority over all baseline methods in both inference speed and computational overhead, achieving a commendable balance between excellent predictive performance and reasonable computational efficiency. These results fully demonstrate the effectiveness of its lightweight and efficient network design. By significantly cutting down computational expenses while maintaining high predictive accuracy, the WDNO model is particularly well-suited for real-time inference tasks in realistic engineering scenarios and serves as a feasible solution for complex multiphysics simulations.

\begin{table}[h]
\renewcommand{\arraystretch}{0.92}
\begin{center}
\resizebox{1\linewidth}{!}{
\begin{tabular}{ccccccccc}
\hline
 & MIONet & U-Net & CODA-NO & GNOT & Transolver & M2NO& WDNO \\
\hline
\rowcolor{blockgray}
Inference Time & 1.21 & 0.78 & 1.52 & 0.30 & 1.18 & 1.25 &0.08 \\
FLOPs          & 211  & 484  & 4855 & 2593 & 184 & 10966 &38  \\
\hline
\end{tabular}
}
\end{center}
\caption{The inference time per forward propagation and floating-point operations (FLOPs) of all models}
\label{fl}
\end{table}

\subsection{One-step Prediction}

In addition to the rollout prediction experiment, we also conducted a one-step prediction experiment to evaluate the model's performance in short-term predictions. The results are shown in the Table \ref{onestep}. WDNO outperformed all other models in terms of RMSE. WDNO performed exceptionally well in the Gravity case, with an RMSE improvement of nearly 60\% compared to U-Net's 0.792 in $\mathbf{V}_\mathbf{x}$ and 0.425 in $\mathbf{V}_\mathbf{y}$. Furthermore, WDNO has a significant advantage in Drug Delivery case, with RMSEs of 0.003 in $\mathbf{C}$ and 0.005 in $\mathbf{P}$ respectively, a substantial improvement compared to Transolver's 0.120 in $\mathbf{C}$ and 0.097 in $\mathbf{P}$. Its performance on the Gravity and Drug Delivery complex shapes datasets was superior to the other two square datasets, indicating that, in the absence of structural priors, its global frequency domain transformation characteristics give WDNO a greater advantage.

In the FRMSE results, WDNO demonstrates superior performance across the entire frequency domain, especially in the high-frequency range, where it outperforms all baselines in all cases. In the Single Bubble case, WDNO achieved excellent FRMSE highs of 0.005 in $\mathbf{T}$, 0.020 in $\mathbf{V}_\mathbf{x}$, and 0.019 in $\mathbf{V}_\mathbf{y}$ across the three fields, compared to M2NO's highs of 0.050 in $\mathbf{T}$, 0.123 in $\mathbf{V}_\mathbf{x}$, and 0.160 in $\mathbf{V}_\mathbf{y}$. In the low-frequency range, WDNO improved the temperature field $\mathbf{T}$ in Gravity case to 0.207 compared to U-Net's 0.807. In the Saturated case, it achieved nearly a 50\% improvement over MIONet's velocity field (4.762 in $\mathbf{V}_\mathbf{x}$ and 5.085 in $\mathbf{V}_\mathbf{y}$, 2.217 in $\mathbf{V}_\mathbf{x}$ and 2.463 in $\mathbf{V}_\mathbf{y}$). This indicates that WDNO achieves prominent optimization for high-frequency feature fitting while fully retaining strong learning capability for low-frequency components, and there is no obvious performance trade-off between low- and high-frequency prediction tasks.

\begin{table*}[t]\scriptsize
\renewcommand{\arraystretch}{1}
\centering
\resizebox{1\linewidth}{!}{
\begin{tabular}{llcccccccccccccc}
\toprule
\multirow{2}{*}{\textbf{Model}} & \multirow{2}{*}{\textbf{Metric}} 
& \multicolumn{3}{c}{\textbf{Subcooled}} 
& \multicolumn{3}{c}{\textbf{Gravity}} 
& \multicolumn{3}{c}{\textbf{Saturated}} 
& \multicolumn{3}{c}{\textbf{Single Bubble}} 
& \multicolumn{2}{c}{\textbf{Drug Delivery}}\\
\cmidrule(lr){3-5} \cmidrule(lr){6-8} \cmidrule(lr){9-11} \cmidrule(lr){12-14} \cmidrule(lr){15-16}
& & $\mathbf{T}$ & $\mathbf{V_x}$ & $\mathbf{V_y}$ 
& $\mathbf{T}$ & $\mathbf{V_x}$ & $\mathbf{V_y}$ 
& $\mathbf{T}$ & $\mathbf{V_x}$ & $\mathbf{V_y}$ 
& $\mathbf{T}$ & $\mathbf{V_x}$ & $\mathbf{V_y}$ & $\mathbf{C}$ & $\mathbf{P}$\\
\hline

\rowcolor{blockgray}
\multirow{4}{*}{MIONet} 
& RMSE & 0.048 & 0.231 & 0.221 & 0.139 & 0.581 & 0.448 & 0.054 & 0.518 & 0.549 & 0.028 & 0.103 & 0.098 & 0.045 & 0.017\\
& FRMSE low & 0.248 & 1.423 & 1.218 & 0.804 & 4.388 & 1.282 & 0.295 & 4.762 & 5.085 & 0.131 & 0.278 & 0.302 & -- & --\\
& FRMSE mid & 0.271 & 1.125 & 1.084 & 0.758 & 3.369 & 2.890 & 0.372 & 3.458 & 3.891 & 0.118 & 0.291 & 0.338 & -- & --\\
& FRMSE high & 0.045 & 0.218 & 0.187 & 0.133 & 0.541 & 0.431 & 0.051 & 0.442 & 0.488 & 0.023 & 0.079 & 0.081 & -- & --\\

\rowcolor{blockgray}
\multirow{4}{*}{U-Net} 
& RMSE & 0.036 & 0.185 & 0.165 & 0.126 & 0.792 & 0.425 & 0.046 & 0.408 & 0.409 & 0.032 & 0.064 & 0.071 & 0.063 & 0.162\\
& FRMSE low & 0.152 & 0.831 & 0.968 & 0.807 & 4.473 & 1.279 & 0.168 & 2.815 & 2.936 & 0.312 & 0.528 & 0.842 & -- & --\\
& FRMSE mid & 0.219 & 0.972 & 0.928 & 0.696 & 3.118 & 2.282 & \textbf{0.315} & 3.295 & 3.517 & 0.285 & 0.594 & 0.563 & -- & --\\
& FRMSE high & 0.039 & 0.174 & 0.142 & 0.120 & 0.504 & 0.392 & 0.044 & 0.368 & 0.346 & 0.027 & 0.055 & 0.062 & -- & --\\

\rowcolor{blockgray}
\multirow{4}{*}{CODA-NO} 
& RMSE & 0.037 & 0.201 & 0.178 & 0.117 & 0.534 & 0.437 & 0.048 & 0.445 & 0.442 & -- & -- & -- & 0.047 & 0.028\\
& FRMSE low & 0.175 & 1.128 & 1.135 & 0.484 & 3.630 & 1.342 & \textbf{0.213} & 3.725 & 3.768 & -- & -- & -- & -- & --\\
& FRMSE mid & 0.208 & 1.042 & 0.985 & 0.676 & 3.430 & 3.036 & 0.341 & 3.486 & 3.542 & -- & -- & -- & -- & --\\
& FRMSE high & 0.036 & 0.192 & 0.157 & 0.112 & 0.495 & 0.416 & 0.047 & 0.405 & 0.384 & -- & -- & -- & -- & --\\

\rowcolor{blockgray}
\multirow{4}{*}{GNOT} 
& RMSE & 0.038 & 0.211 & 0.217 & 0.118 & 0.661 & 0.488 & 0.051 & 0.475 & 0.491 & 0.010 & 0.069 & 0.070 & 0.055 & 0.020\\
& FRMSE low & 0.185 & 1.285 & 1.698 & 0.696 & 5.781 & 3.543 & 0.271 & 3.985 & 4.428 & 0.129 & 0.728 & 0.967 & -- & --\\
& FRMSE mid & 0.221 & 0.988 & 1.023 & 0.664 & 3.630 & 3.461 & 0.362 & 3.685 & 3.792 & 0.078 & 0.741 & 0.512 & -- & --\\
& FRMSE high & 0.037 & 0.183 & 0.195 & 0.113 & 0.611 & 0.449 & 0.048 & 0.427 & 0.419 & 0.009 & 0.063 & 0.057 & -- & --\\

\rowcolor{blockgray}
\multirow{4}{*}{Transolver} & RMSE& 0.038 & 0.188 & 0.162 & 0.121 & 0.539 & 0.435 & 0.049 & 0.464 & 0.464 & 0.012 & 0.074 & 0.082 & 0.120 & 0.097\\
& FRMSE low & 0.216 & 1.002 & 0.957 & 0.529 & 3.244 & 1.824 & 0.264 & 3.896 & 3.961 & 0.153 & 0.992 & 1.289 & -- & --\\
& FRMSE mid & 0.209 & 0.970 & 0.887 & 0.670 & 3.414 & 2.812 & 0.354 & 3.535 & 3.593 & 0.101 & 0.725 & 0.610 & -- & --\\
& FRMSE high & 0.035 & 0.176 & 0.149 & 0.117 & 0.507 & 0.417 & 0.046 & 0.414 & 0.413 & 0.010 & 0.064 & 0.070 & -- & --\\

\rowcolor{blockgray}
\multirow{4}{*}{M2NO} & RMSE& 0.034 & 0.167 & 0.146 & 0.118 & 0.498 & 0.390 & 0.048 & 0.355 & 0.367 & 0.050 & 0.123 & 0.160 & 0.048 & 0.030\\
& FRMSE low & 0.118 & 0.655 & 0.696 & 0.451 & 3.048 & 1.352 & 0.223 & 2.500 & 2.718 & 0.197 & 0.568 & 0.717 & -- & --\\
& FRMSE mid & 0.195 & 0.887 & 0.783 & 0.622 & 2.948 & 2.187 & 0.349 & 3.820 & 3.878 & 0.182 & 0.391 & 0.701 & -- & --\\
& FRMSE high & 0.032 & 0.159 & 0.129 & 0.114 & 0.461 & 0.400 & 0.045 & 0.364 & 0.374 & 0.045 & 0.107 & 0.121 & -- & --\\

\rowcolor{oursblue}
\multirow{4}{*}{\textbf{WDNO}} 
& RMSE & \textbf{0.033} & \textbf{0.171} & \textbf{0.142} 
& \textbf{0.041} & \textbf{0.184} & \textbf{0.164} 
& \textbf{0.046} & \textbf{0.361} & \textbf{0.361} 
& \textbf{0.005} & \textbf{0.020} & \textbf{0.019} & \textbf{0.003} & \textbf{0.005}\\
& FRMSE low & \textbf{0.119} & \textbf{0.763} & \textbf{0.694}
& \textbf{0.207} & \textbf{1.056} & \textbf{1.122}
& 0.223 & \textbf{2.217} & \textbf{2.463}
& \textbf{0.032} & \textbf{0.183} & \textbf{0.256} & -- & --\\
& FRMSE mid & \textbf{0.180} & \textbf{0.857} & \textbf{0.773}
& \textbf{0.289} & \textbf{1.258} & \textbf{1.279}
& 0.321 & \textbf{2.843} & \textbf{2.980}
& \textbf{0.030} & \textbf{0.132} & \textbf{0.139} & -- & --\\
& FRMSE high & \textbf{0.031} & \textbf{0.159} & \textbf{0.128}
& \textbf{0.040} & \textbf{0.175} & \textbf{0.151}
& \textbf{0.044} & \textbf{0.330} & \textbf{0.333}
& \textbf{0.004} & \textbf{0.017} & \textbf{0.016} & -- & --\\
\bottomrule
\end{tabular}}
\caption{The Main Results of Multi-Field one-step Prediction.
The CODA-NO model failed to produce effective results for the Single Bubble case due to unstable training. The best results are shown in bold.}
\label{onestep}
\end{table*}

\subsection{Ablation Study}
\begin{figure}[!htbp]
  \centering
  \includegraphics[width=0.418\textwidth]{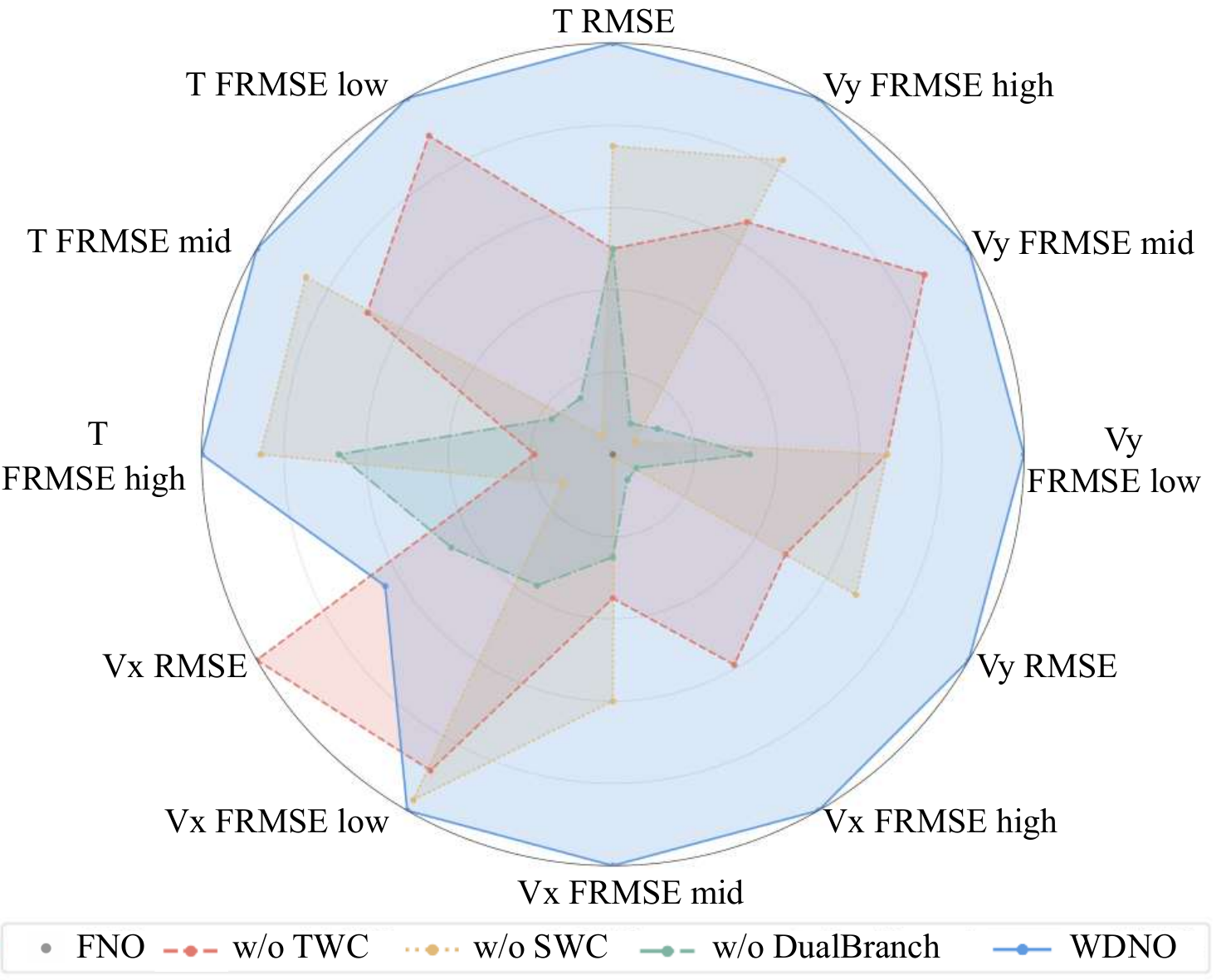}
  \caption{The prediction results on Subcooled case. The outer edge represents the minimum loss.}
  \label{fig:rar}
\end{figure}

The radar chart (Fig. \ref{fig:rar}) intuitively visualizes the module ablation results, with all metrics normalized using the maximum and minimum values of all models. As observed, w/o TWC, SWC, and DualBranch, the performance across the 12 metrics varies, and all are superior to the base FNO. This solidly substantiates that the three modules operate as standalone functional components and feature complementary strengths.

\section{Limitations}
While WDNO delivers excellent performance for spatio-temporal multi-scale long-term prediction, it has two main limitations. First, it employs fixed Haar wavelet kernels for decomposition without adaptive weight tuning. While fixed Haar wavelet kernels help improve computational efficiency significantly, their static weights fail to dynamically balance the contributions of low-frequency global trends and high-frequency local details in complex physical scenarios with strong spatiotemporal scale coupling, such as Gravity case. To address this issue, future work will focus on developing a learnable dynamic weight adjustment module to enhance the model’s adaptive capability while preserving its inherent computational efficiency.

Second, WDNO adopts only Haar wavelets as the unified decomposition basis. Highly sensitive to signal discontinuities, Haar wavelets are well-suited for non-stationary physical processes with strong discontinuities, e.g., bubble dynamics involving gas-liquid interfaces, which justifies the adoption of Haar wavelets in this work. Nevertheless, the exclusive use of Haar wavelets restricts the model’s adaptability to diverse physical field patterns, including smooth flow fields, strong discontinuities and evolving complex interfaces. On this basis, optimal wavelet bases will be tailored for different types of physical prediction tasks to further elevate the model’s universal applicability.

\section{Dataset}
\textbf{BubbleML} Developed explicitly for machine learning (ML) research, BubbleML is a comprehensive multiphysics dataset focused on phase-change boiling. Its primary objective is to alleviate the dearth of high-caliber datasets that hinder ML-based investigations into phase-change processes. A key limitation of existing experimental datasets lies in their restricted applicability and insufficient ground truth annotations; meanwhile, most contemporary scientific machine learning (SciML) datasets are confined to single-phase physical phenomena, lacking the ability to encapsulate the complex features inherent to multi-phase phase changes.
Leveraging physics-informed simulation techniques via the Flash-X framework, the dataset incorporates 79 distinct sets of simulation results, encompassing a wide spectrum of boiling configurations—nucleate pool boiling, flow boiling, and subcooled boiling among them. It covers an expansive parameter space, ranging from terrestrial to microgravity gravitational environments, alongside varied flow velocities, subcooling degrees, and wall superheat magnitudes. Essential physical quantities, including fluid velocity, temperature distribution, pressure fields, and the signed distance function that delineates liquid-vapor interfaces, are archived in the HDF5 format. To ensure its reliability and precision, the dataset has been subjected to rigorous experimental validation procedures.

To facilitate its application in ML research, two distinct types of benchmark tasks have been constructed for this dataset: one focusing on optical flow analysis to characterize bubble dynamics, and the other centered on SciML-based modeling for thermal dynamic predictions. Aligned with the FAIR (Findable, Accessible, Interoperable, Reusable) guidelines, the dataset is freely accessible to the research community via GitHub and Zenodo. This initiative addresses the existing gap in ML datasets for multi-phase phase-change studies, providing researchers with high-fidelity, diverse data resources and standardized evaluation benchmarks to advance ML-driven explorations of phase-change phenomena.

For the experimental validation portion of the study, four specific cases were selected: Subcooled Pool Boiling, Flow Boiling under Gravitational Conditions, Saturated Pool Boiling, and Single Bubble experiments.

\begin{figure*}[t]
    \centering
    \includegraphics[width=0.8\textwidth]{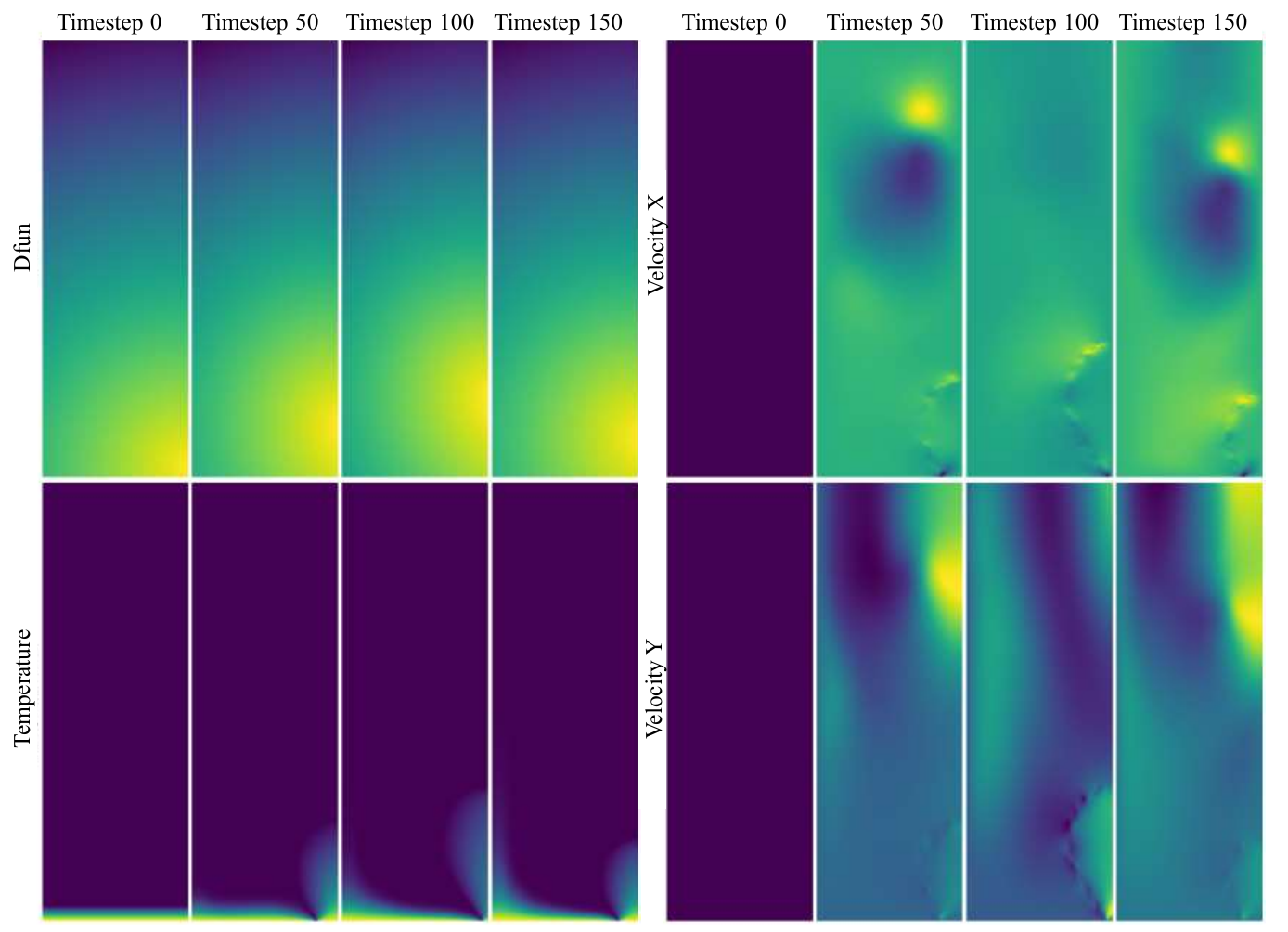}
    \caption{
    Single Bubble Visualization
    }
    \label{Single Bubble}
\end{figure*}

\textbf{Single Bubble} Within a confined spatial environment, an isolated bubble undergoes the entire sequence of its life cycle—beginning with nucleation and subsequent growth, followed by detachment from the heating element located at the bottom, and concluding with upward migration—without being affected by the presence of other bubbles. As the fundamental building block for exploring complex boiling mechanisms, this single-bubble system plays a crucial role in two key aspects: first, it facilitates the extraction of essential multiphase flow regulations, including the associations between bubble detachment rate, ascending speed, and heat input; second, it provides a basis for the adjustment and validation of machine learning models, such as simplified tasks involving the dynamic tracking of bubble movement. Below is a comprehensive breakdown of the dataset’s specifications and core configuration parameters.
The simulation scenario is composed of a single 2D model (designated as PB-Single Bubble). Key specifications of this model include a physical domain measuring 4.2 mm × 6.3 mm, a spatial resolution of 192 × 288 pixels, and a time increment of $\Delta t = 0.5$—this time step translates to a physical duration of 0.004 seconds. To ensure complete coverage of the single bubble’s life cycle, from initial nucleation to final ascent, the simulation incorporates a total of 500 time steps, resulting in an overall data volume of 0.5 GB.

All core parameters of the simulation are set uniformly, as detailed below: The working fluid selected for the simulation is FC-72 (perfluorohexane), a substance characterized by its electrical insulating properties and suitability for low-temperature cooling operations in the 50–100 °C range. Gravitational conditions are standardized to terrestrial gravity, with a/g = 1. The temperature of the heating surface is calibrated to lie within the nucleate boiling range specific to FC-72; this configuration guarantees the stable formation of bubbles while preventing any disruptions caused by condensation or bubble coalescence.

The shape of the bubble and its interface are precisely quantified using the signed distance function (SDF), represented by $\phi$. In this framework, a positive value of $\phi$ ($\phi \textgreater 0$) denotes the vapor phase, whereas a negative value ($\phi \textless 0$) corresponds to the liquid phase. The initial bubble is placed at a predetermined nucleation point on the heater surface, taking the form of a small sphere with an initial velocity of 0. Notably, the simulation is free from initial flow disturbances or pressure gradient effects, which allows for the unobstructed capture of the intrinsic physical dynamics governing the behavior of a single bubble.

\begin{figure*}[t]
    \centering
    \includegraphics[width=0.6\textwidth]{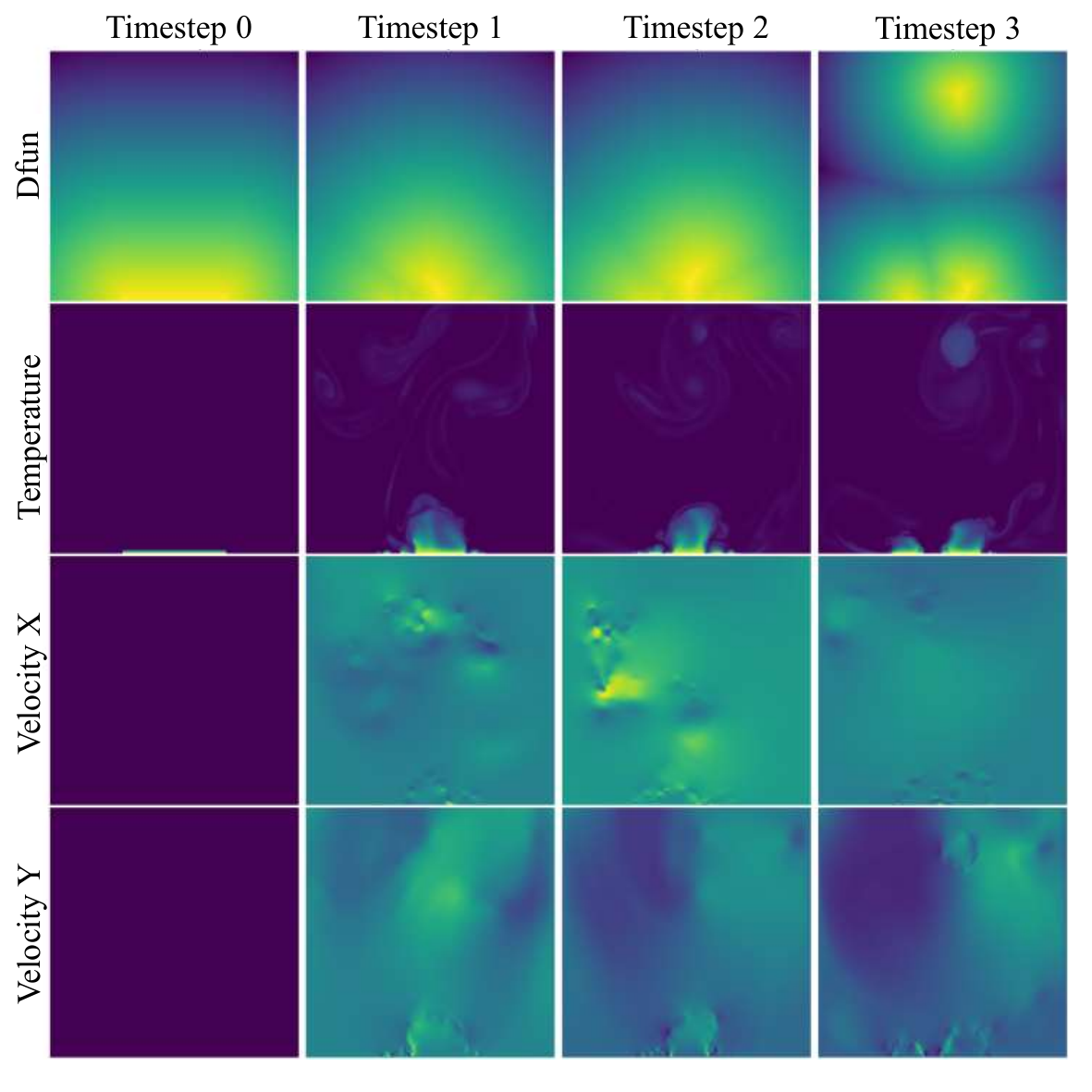}
    \caption{
    Subcooled Pool Boiling
    }
    \label{Subcooled Pool}
\end{figure*}

\textbf{Subcooled Pool Boiling} A defining feature of subcooled boiling lies in the fact that the overall temperature of the liquid phase remains lower than its saturation temperature. When bubbles initiate (nucleate) and expand on the surface of the bottom-mounted heater, they undergo condensation upon contact with the surrounding subcooled liquid—this process demonstrates dynamic behaviors characterized by the simultaneous occurrence of evaporation and condensation. This particular boiling phenomenon is mainly applied in practical scenarios that demand efficient bubble-driven heat transfer, such as the high-performance cooling of electronic components and low-temperature thermal management systems.
The detailed specifications and key parameters of the corresponding dataset are outlined as follows:

- PB-Subcooled: This subset comprises 10 independent simulation runs, with a physical computational domain measuring 8.4 mm × 8.4 mm and a spatial resolution of 384 × 384 pixels. The time step is set to $\Delta t = 1$, which corresponds to a physical time interval of 0.008 seconds, and the total number of time steps reaches 200, resulting in a total data volume of 10.3 GB.

- PB-Subcooled 0.1: Consisting of 15 simulation trials, this subset shares the same physical domain dimensions and spatial resolution as the PB-Subcooled dataset. However, it adopts a smaller time step of $\Delta t = 0.1$ (equivalent to a physical time of 0.0008 seconds) and includes 2000 time steps in total, leading to a larger data volume of 155.1 GB.

The core parameters for all simulations are uniformly configured as follows: The working fluid employed is FC-72 (perfluorohexane), a substance with excellent electrical insulation capabilities, making it well-suited for low-temperature cooling applications within the range of 50–100 °C. The subcooling effect is realized by regulating the temperature difference between the bulk liquid temperature ($\text{T}_{bulk}$) and the saturation temperature of FC-72 (around 58 °C), which ensures a distinct condensation effect during the boiling process. The heater temperature is adjusted between 60–110 °C, a range that covers the nucleate boiling regime and prevents the occurrence of critical heat flux (CHF). Gravitational conditions are set to the standard terrestrial gravity (a/g = 1) by default. The liquid-vapor interface is monitored and tracked using the Signed Distance Function (SDF), denoted as $\phi$—where $\phi > 0$ indicates the vapor phase and $\phi < 0$ denotes the liquid phase. Additionally, the boundary conditions of the simulation domain are fixed to eliminate any extraneous disturbances, thereby focusing the analysis on the impacts of subcooling degree on bubble dynamics and heat transfer flux.

\begin{figure*}[t]
    \centering
    \includegraphics[width=0.6\textwidth]{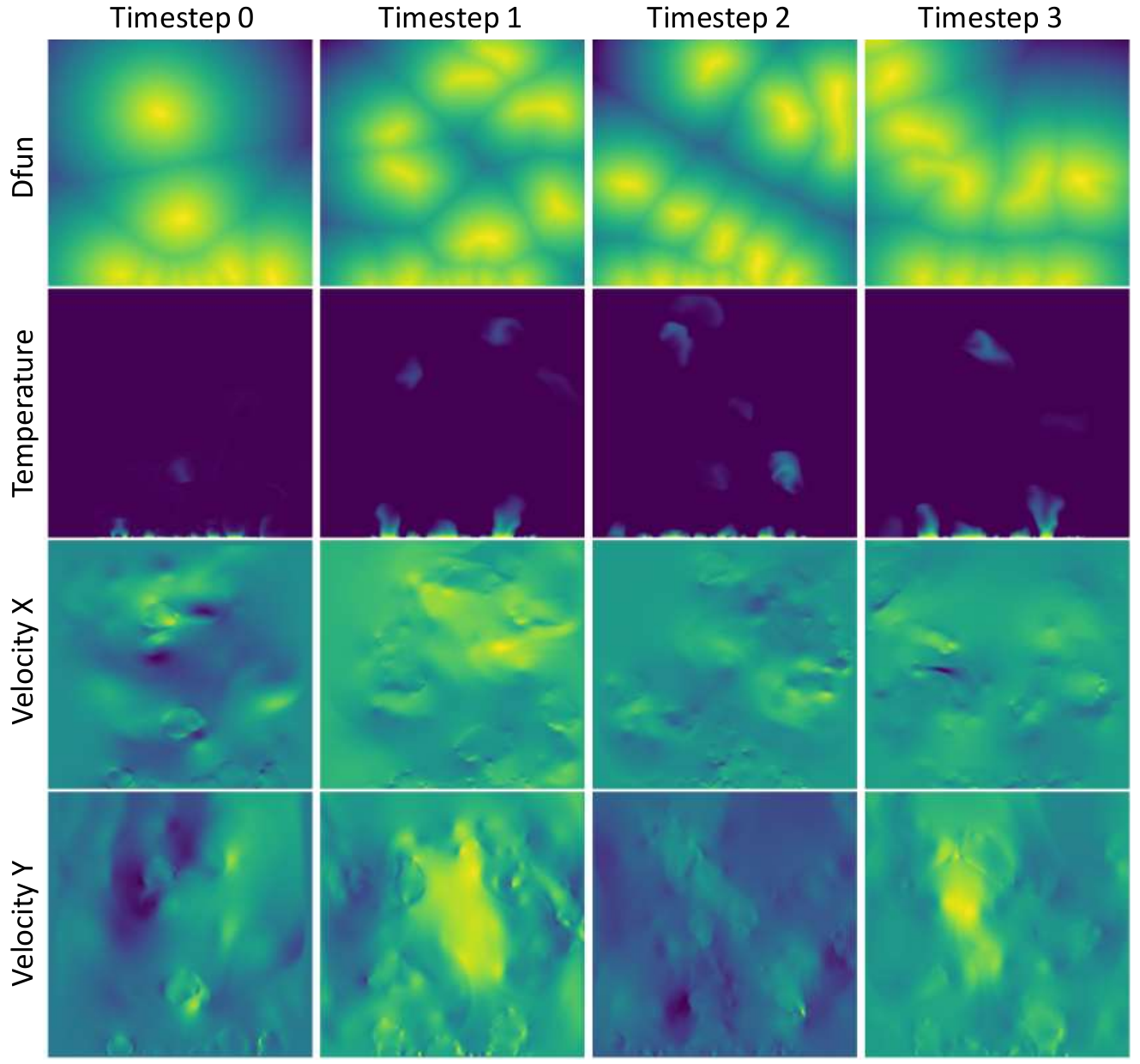}
    \caption{
    Saturated Pool Boiling
    }
    \label{Saturated Pool}
\end{figure*}

\textbf{Saturated Pool Boiling} A defining physical trait of saturated pool boiling lies in the equality between the liquid’s bulk temperature and its saturation temperature, a state where the liquid and vapor phases maintain thermal equilibrium. Upon nucleation and growth on the heater surface, bubbles do not experience condensation-driven contraction; instead, their detachment and upward movement are solely driven by buoyant forces. This particular scenario is pivotal to investigating the shift from nucleate boiling to critical heat flux, and it holds broad significance for addressing boiling heat transfer challenges in non-subcooled environments—such as those encountered in nuclear waste thermal management and electronic component heat dissipation.
The saturated pool boiling process is examined through 13 distinct, independent simulations. Heater temperatures are uniformly spaced at 5°C increments, ranging from 60°C to 120°C, which encompasses the transition of boiling flow patterns from bubbly flow to slug flow. The simulated physical domain measures 11.2mm × 11.2mm, with a spatial resolution of 512 × 512 pixels. For temporal discretization, a time step $\Delta t = 1$ (dimensionless time, equivalent to 0.008 seconds of physical time) is adopted, and each simulation incorporates 200 time steps—an interval sufficient to capture the full lifecycle of multiple bubble nucleation-detachment cycles. The total dataset amounts to 24.2GB, all of which is stored in the HDF5 file format.

FC-72 (perfluorohexane) serves as the working fluid in these simulations. This fluid exhibits electrical insulation capabilities, making it well-suited for low-temperature cooling applications within the 50–100°C range. Simulations are conducted under terrestrial gravity conditions (no low-gravity configurations are implemented), with a primary focus on capturing the fundamental physical principles governing saturated boiling in standard environments.

\begin{figure*}[t]
    \centering
    \includegraphics[width=\textwidth]{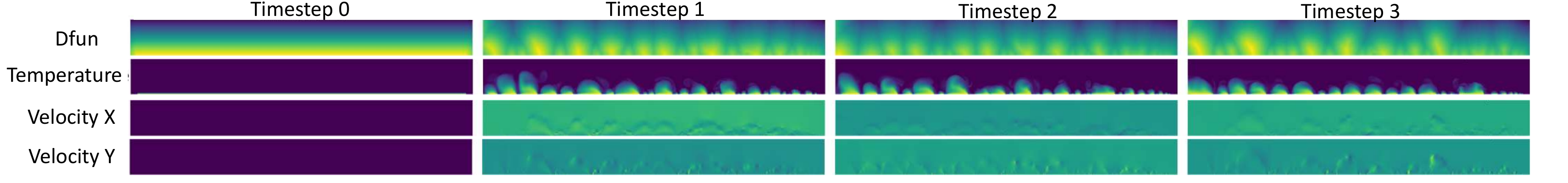}
    \caption{
    Flow Boiling Gravity
    }
    \label{Flow Boiling}
\end{figure*}

\textbf{Flow Boiling Gravity} The Flow Boiling Gravity experimental system is engineered to investigate how gravitational forces influence bubble dynamic behaviors, the evolutionary characteristics of flow regimes, and heat transfer flux. This is accomplished by adjusting gravitational environments, which span from standard terrestrial gravity conditions ($a/g=1$) to the microgravity environment encountered aboard the International Space Station ($a/g=10^{-4}$). Its primary applications are focused on practical fields such as space exploration missions and thermal management systems operating under extreme environmental conditions. Below is a comprehensive breakdown of the specifications and key parameters associated with its dataset.

- FB-Gravity Dataset: Comprising 6 independent simulation runs, this dataset features a physical computational domain measuring 35 mm × 3.5 mm, with a spatial resolution of 1600 × 160 pixels. The simulation employs a time increment of $\Delta t=1$, which translates to an actual physical time interval of 0.008 seconds. Each simulation consists of 200 discrete time steps, resulting in a total data volume of 10.9 GB.

- FB-Gravity 0.1 Dataset: This dataset also includes 6 simulation runs, maintaining the same physical domain dimensions and spatial resolution as the FB-Gravity dataset. However, it utilizes a smaller time increment of $\Delta t=0.1$ (corresponding to a physical time of 0.0008 seconds) and a greater number of time steps (2000 in total), leading to a larger overall data volume of 108.6 GB.

A set of consistent core parameters is applied across all simulations: The working fluid selected is FC-72 (perfluorohexane), a substance characterized by its electrical insulation capabilities and suitability for low-temperature cooling applications within the range of 50–100 $^\circ C$. Gravitational scaling is implemented through the Froude number, with $\text{Fr}^{-2}=a/g$, encompassing a gradient from $10^{-4}$ to 1. The inlet flow velocity is fixed at the threshold velocity required to initiate bubbly flow during flow boiling, while the heater temperature is controlled within the nucleate boiling range of FC-72 (80–100 $^\circ C$). This parameter configuration ensures stable bubble nucleation while preventing the occurrence of critical heat flux, which would disrupt the experimental conditions.

\textbf{Multiphysics Drug Delivery} The respiratory drug delivery simulation dataset is constructed to quantitatively investigate the coupled multiphysics evolution of respiratory airflow and aerosol drug concentration transport inside human upper airway anatomical structures. This dataset characterizes airflow velocity, pressure distribution and scalar drug advection-diffusion transport under varied inhalation boundary conditions and initial drug concentration loads, supporting the development of physics-embedded graph network simulators for medical respiratory thermal-fluid analysis. Its primary applications cover pulmonary drug inhalation therapy simulation, personalized respiratory treatment optimization, and unstructured mesh-based multiphysics long-term prediction research. Below is a comprehensive breakdown of the specifications and key parameters associated with its dataset.

\begin{figure*}[t]
    \centering
    \includegraphics[width=1\textwidth]{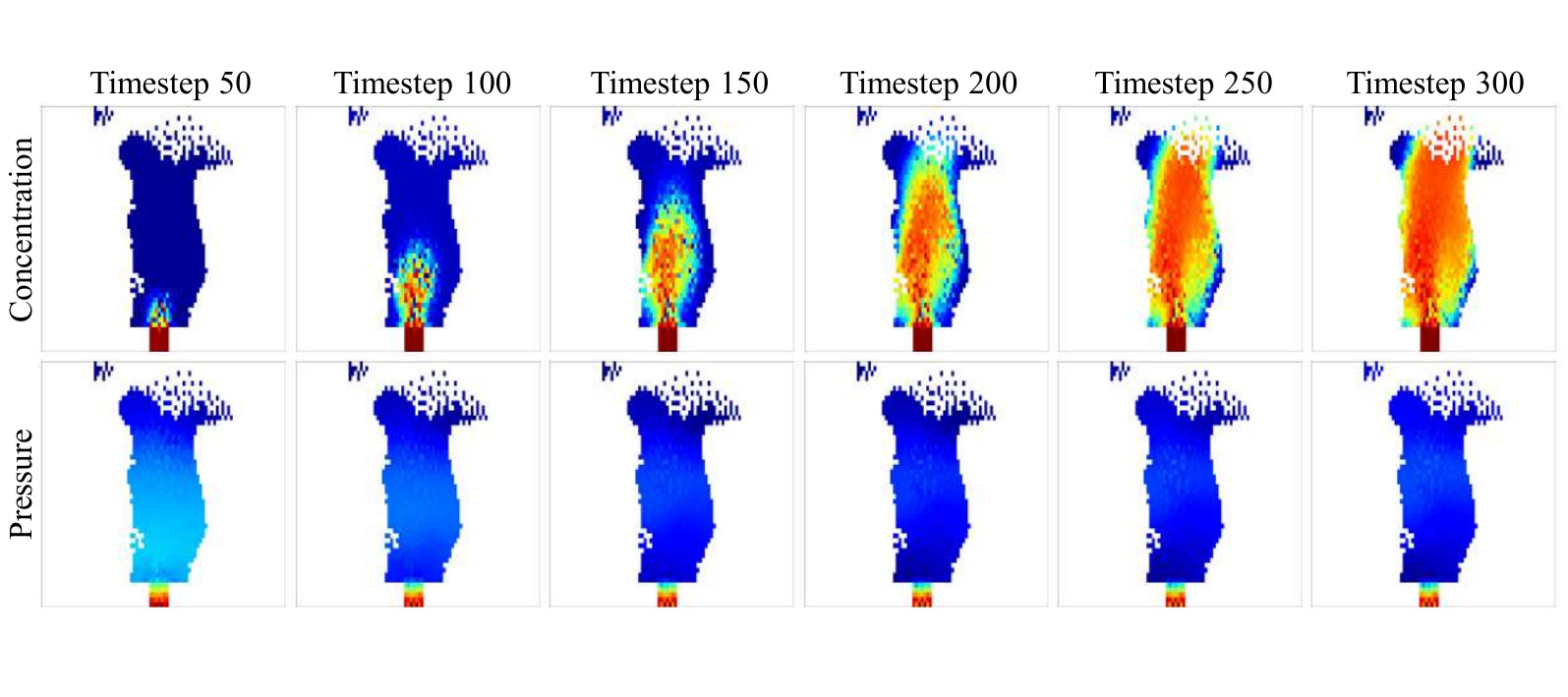}
    \caption{
    Drug Delivery
    }
    \label{drug}
\end{figure*}

Respiratory Drug Delivery Dataset: Comprising 144 independent simulation trajectories, this dataset features a 3D tetrahedral mesh computational domain reconstructed from simplified human upper respiratory tract geometry equipped with an oral drug delivery mouthpiece. The simulation employs a time increment of $\Delta t=0.005$, which corresponds to an actual physical time interval of 0.005 seconds. Each simulation trajectory consists of 400 discrete time steps, covering a complete inhalation cycle of 2 seconds, with 120 trajectories allocated for model training and 24 for testing.
A set of consistent core parameters is applied across all simulation trajectories: The governing physical laws are the incompressible Navier-Stokes equations for fluid airflow and advection-diffusion equations for scalar drug concentration transport. The spatial mesh adopts 3D tetrahedron unstructured grids to fit complex anatomical airway geometry, with node features including spatial coordinates, node type one-hot encoding, flow velocity, fluid pressure, and drug concentration scalar field. Inlet airflow boundary conditions and initial aerosol drug concentration distributions are set to multiple gradient levels to diversify physical flow states. The simulation strictly follows mass conservation and fluid divergence-free physical constraints, avoiding flow separation artifacts and unphysical mass loss, which provides high-fidelity ground truth data for evaluating the physical consistency and long-term prediction stability of multiphysics deep learning simulators.

\end{document}